%% file: main.tex
\documentclass[11pt, a4paper, copyright, gdm]{google}

\usepackage[authoryear, sort&compress, round]{natbib}
\input{math_commands.tex}

\usepackage{enumitem} %
\usepackage{hyperref}
\usepackage{url}

\hypersetup{
  breaklinks=true,
}

\usepackage{graphicx} %
\usepackage{multirow} %

\usepackage{booktabs}
\usepackage[raggedrightboxes]{ragged2e}
\usepackage{longtable}  %

\usepackage{cleveref,xspace}
\usepackage{graphicx}
\usepackage{caption, subcaption}
\usepackage[version=4]{mhchem}  %
\usepackage{siunitx}
\usepackage[svgnames,table]{xcolor} %
\usepackage{wrapfig} 
\usepackage{enumitem}

\usepackage{fancyvrb}
\usepackage{fvextra}

\usepackage{tcolorbox}       %
\tcbuselibrary{listings,breakable}
\usepackage{listings}        %

\definecolor{examplebg}{named}{Gainsboro}

\newtcolorbox{modelanswer}[2][]{
  colback=#2!5!white,
  colframe=#2!75!black,
  fonttitle=\bfseries,
  title=#1,
  breakable,         %
  boxsep=1mm,        %
  toptitle=1mm,
  bottomtitle=1mm,
}

\newtcolorbox{examplebox}{
  title=Example Answer,
  colback=examplebg,
  colframe=black!75!white,
  boxrule=0.5pt,
  arc=2mm,
  breakable, 
}

\definecolor{rankonecolor}{HTML}{6ABE7A}   %
\definecolor{ranktwocolor}{HTML}{92D08E}   %
\definecolor{rankthreecolor}{HTML}{C5E0B4}  %
\definecolor{rankfourcolor}{HTML}{D8E7A9}  %
\definecolor{rankfivecolor}{HTML}{FFEB84}   %
\definecolor{ranksixcolor}{HTML}{FFD966}   %
\definecolor{ranksevencolor}{HTML}{FFA500}  %

\newcommand{\rankone}{\cellcolor{rankonecolor}}
\newcommand{\ranktwo}{\cellcolor{ranktwocolor}}
\newcommand{\rankthree}{\cellcolor{rankthreecolor}}
\newcommand{\rankfour}{\cellcolor{rankfourcolor}}
\newcommand{\rankfive}{\cellcolor{rankfivecolor}}
\newcommand{\ranksix}{\cellcolor{ranksixcolor}}
\newcommand{\rankseven}{\cellcolor{ranksevencolor}}

\newcommand{\name}{CLINB\xspace}
\newcommand{\geminiflash}{Gemini 2.5 Flash\xspace}
\newcommand{\geminipro}{Gemini 2.5 Pro\xspace}
\newcommand{\hybrid}{Hybrid\xspace}
\newcommand{\opus}{Claude Opus 4.1\xspace}
\newcommand{\sonnet}{Claude Sonnet 4\xspace}
\newcommand{\gpt}{GPT-5\xspace}
\newcommand{\othree}{OpenAI o3\xspace}

\newcommand{\editor}{Editor}

\uselogo{} 

\title{\name: A Climate Intelligence Benchmark for Foundational Models}

\correspondingauthor{michellechen@google.com. Authors names in random order.}

\reportnumber{0001} %

\renewcommand{\today}{2025-10-25}

\author[1]{Michelle Chen Huebscher}
\author[2]{Katharine Mach}
\author[1]{Aleksandar Stani\'{c}}
\author[1,3]{Markus Leippold}
\author[1]{Ben Gaiarin}
\author[4]{Zeke Hausfather}
\author[ ]{Elisa Rawat}
\author[5]{Erich Fischer}
\author[1]{Massimiliano Ciaramita}
\author[6]{Joeri Rogelj}
\author[1]{Christian Buck}
\author[1]{Lierni Sestorain Saralegui}
\author[5]{Reto Knutti}

\affil[1]{\thepa{}{}}
\affil[2]{University of Miami}
\affil[3]{University of Zurich}
\affil[4]{Stripe}
\affil[5]{ETH Zurich}
\affil[6]{Imperial College London}

\begin{abstract}
\input{sections/0-abstract}
\end{abstract}

\begin{document}

\maketitle

\section{Introduction}
\input{sections/1-introduction}

\section{\name Data}
\input{sections/2-dataset}

\section{Experiments and Analysis}
\input{sections/4-results}

\section{Related Work}
\input{sections/5-related-work}

\section{Conclusion}
\input{sections/conclusion}

\section*{Acknowledgments}
We would like to thank Peter Battaglia, Taylan Cemgil, Pushmeet Kohli, Andrew McCallum and Jeff Sternberg for their feedback and support. Special thanks to Frederic Giuli and Fred Lipschultz.

\bibliography{main}

\newpage
\appendix

\input{sections/20-appendix}

\end{document}

%% file: math_commands.tex
\usepackage{amsmath,amsfonts,bm}

\def\eqref#1{equation~\ref{#1}}

\def\1{\bm{1}}

\DeclareMathAlphabet{\mathsfit}{\encodingdefault}{\sfdefault}{m}{sl}
\SetMathAlphabet{\mathsfit}{bold}{\encodingdefault}{\sfdefault}{bx}{n}

%% file: sections/0-abstract.tex
Evaluating how Large Language Models (LLMs) handle complex, specialized knowledge remains a critical challenge. We address this through the lens of climate change by introducing \name, a benchmark that assesses models on open-ended, grounded, multimodal question answering tasks with clear requirements for knowledge quality and evidential support. \name relies on a dataset of real users' questions and evaluation rubrics curated by leading climate scientists. We implement and validate a model-based evaluation process and evaluate several frontier models. Our findings reveal a critical dichotomy. Frontier models demonstrate remarkable knowledge synthesis capabilities, often exhibiting PhD-level understanding and presentation quality. They outperform ``hybrid" answers curated by domain experts assisted by weaker models. However, this performance is countered by failures in grounding. The quality of evidence varies, with substantial hallucination rates for references and images. We argue that bridging this gap between knowledge synthesis and verifiable attribution is essential for the deployment of AI in scientific workflows and that reliable, interpretable benchmarks like \name are needed to progress towards building trustworthy AI systems.

%% file: sections/1-introduction.tex
A secure path towards Artificial General Intelligence (AGI) depends on the ability to effectively assess AI systems, to foster and track progress, as well as alignment with desired objectives~\citep{Russell2019HumanCompatible}. Significant research is dedicated to evaluation, and performance on popular benchmarks like Chatbot Arena~\citep{chatbotareana} has become a primary driver of development.
The rapid advancement of AI necessitates increasingly challenging benchmarks, which in turn demand technical sophistication and domain knowledge~\citep{gpqa,hle}. Current 'hard benchmarks' are objectively difficult and beneficial to model improvement. 
However, the current approach to building these benchmarks has limitations. 
Firstly, tasks (prompts) often consist of esoteric puzzles and trivia, largely obscure to those outside a niche specialty. 
Secondly, to enable algorithmic verifiability, responses are typically limited to \emph{closed form} tasks \citep{gpqa}; e.g., short text, or numeric, answers and multiple-choice questions \citep{Dinh2024SciExBLA, justen2025llmsoutperformexpertschallenging,hle}. These  represent only a fraction of real-world applications of Generative AI, and the challenges that arise from its use.

We focus on open ended, \emph{generative}, question answering tasks and collaborative Human-AI environments. 
To understand, and improve, how foundational models handle climate change information we develop a benchmark, \name (\textbf{Cl}imate \textbf{In}telligence \textbf{B}enchmark), consisting of a dataset of real users' questions \citep{vaghefi2023chatclimate} and a rubrics-based evaluation pipeline.
Climate change is a broad, complex and hotly debated topic, ultimately rooted in extensive scientific knowledge spanning multiple disciplines, from physical to social sciences. Moreover, the topic draws upon decades of institutional knowledge and best practices. \name's questions require research, evidence-assessment and synthesis skills. These fall under Long-Form Question Answering (LFQA) \citep{fan2019eli5,arora2025healthbench}, where ensuring faithfulness and reliable attribution remain core challenges \citep{ji2023survey}.
For each question, multiple answers are curated by experts with advanced academic knowledge on the subject. Answers consist of free-form text, may contain visual content, and must be supported by robust evidence. 
All answers are evaluated by the experts in side-by-side experiments and the human feedback is used to compile grading rubrics which are validated by climate scientists (among the co-authors of the paper) with experience as leads in main institutional reports.\footnote{E.g., IPCC (\url{https://www.ipcc.ch/}) and NCA (\url{https://toolkit.climate.gov/NCA5}).} 
The rubrics are then used for automatic assessment by a model-based \emph{autorater}. 
This utilizes the "LLM-as-a-Judge" paradigm \citep{zheng2023judging,arora2025healthbench}, employing criteria-based approaches shown to improve alignment with human experts \citep{kim2024prometheus}. 
The whole data creation process is performed in an AI-assisted tool.

We perform an empirical assessment of several frontier models. 
Our main contributions are:

\paragraph{A New Expert-Grounded Benchmark for Scientific AI}
We introduce \name, a benchmark for model-based evaluation of frontier models on complex, multimodal scientific communication. Its core is a new dataset of real-world climate questions paired with \textit{data-driven, question-specific evaluation rubrics}, curated and validated by leading climate scientists through a novel three-phase, human-in-the-loop process.\footnote{We make the system prompts, questions, and model judge prompts available at \url{https://www.kaggle.com/datasets/deepmind/clinb-questions}.}

\paragraph{PhD-Level Synthesis vs. Attribution Failures}
Frontier models demonstrate remarkable knowledge synthesis, often exhibiting a \textit{PhD-level understanding}. However, this performance masks a critical inadequacy in grounding. We report substantial hallucination rates for references ({10\% to 25\%}) and even more failures for images ({50\% to 80\% in certain settings}), exposing a major gap between synthesis and verifiable attribution.

\paragraph{Insights into Human-AI Collaboration Dynamics}
Autonomous frontier models \textit{surpass 'hybrid' answers} (curated by experts using weaker AI assistance), revealing the assisting model's capability—not human oversight—as the primary bottleneck. Counter-intuitively, highly motivated non-specialists (our 'Advocates') who deeply engage with AI tools can produce \textit{higher-quality answers than domain experts} who engage less with AI during answer curation.

\paragraph{A Validated Methodology for Scalable Oversight}
We validate a rigorous, rubric-based autorater. Ablation studies demonstrate that structured prompts and \textit{automated evidence-checking} are essential for mitigating inherent LLM judge biases. This process is  hampered by inaccessible sources (up to {50\%}). Furthermore, we identify evaluation challenges, including \textit{model familiarity bias} in human raters and the limitations of \textit{rubrics to generalize} across models.

%% file: sections/2-dataset.tex
\begin{figure*}
    \centering
    \includegraphics[width=0.99\linewidth]{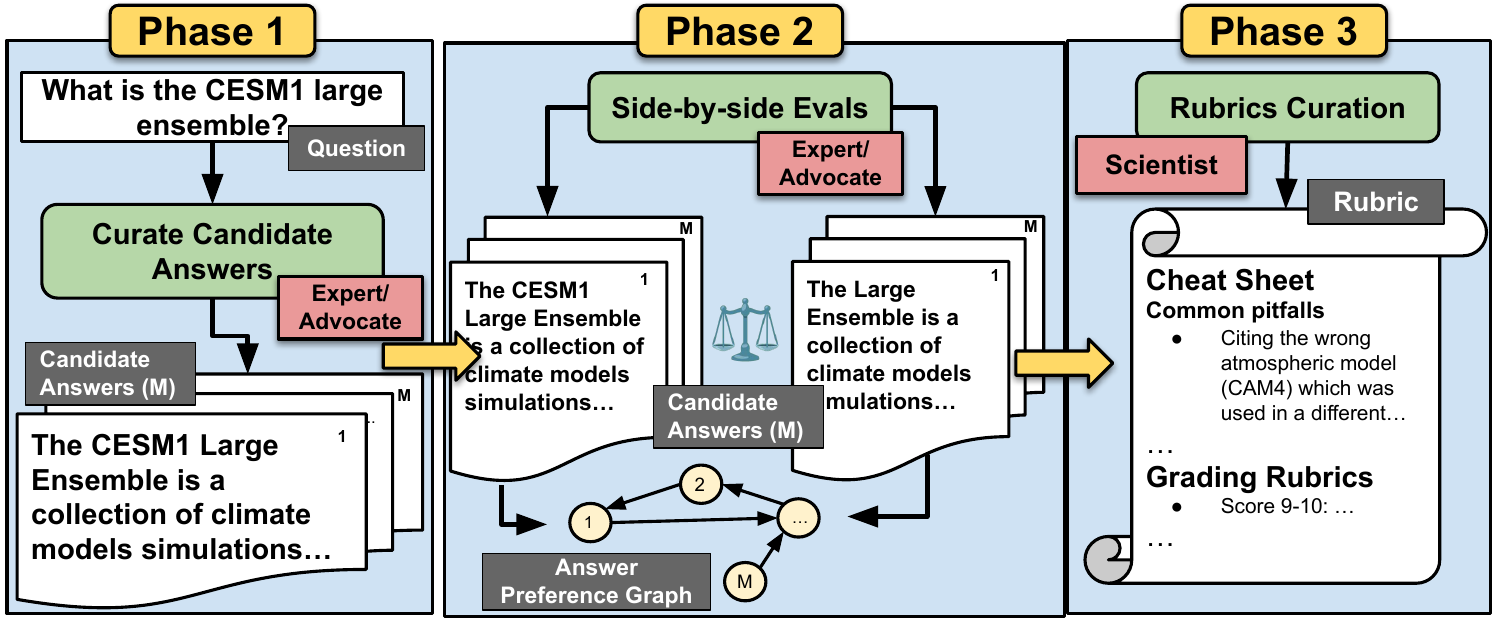}
    \caption{The multi-step, human-in-the-loop process to construct the \name dataset.}
    \label{fig:placeholder}
\end{figure*}

Here we explain the data creation process underlying the \name benchmark. The data consist of human-curated questions, answers, pairwise preferences over pairs of answers and finally question-specific answer-grading instructions: the rubrics. The data is curated by three groups of human experts: Advocates, Experts and Scientists, in three phases illustrated in \Cref{fig:placeholder}.

\subsection{Humans in the Loop}
Human expertise is critical for assessing AI in knowledge intensive tasks, and is a scarce resource\citep{gpqa,hle}.
We intentionally organized a group among the authors, which we call the \emph{Scientists}, to validate the study. This group consists of six academics. Five are climate scientists with extensive expertise and multiple lead roles in IPCC and NCA reports, and the sixth is a leading expert in Climate Finance.

The Scientists curated parts of the data and oversaw the scientific validity of the work, but do not have the capacity to produce sufficient amounts of data.
To scale human data collection 
we formed a pool of 40 experts by directly recruiting active academics (mostly PhD students and postdocs) with the necessary domain expertise. We call this group the \emph{Experts}. To increase diversity and exploration we further recruited 17 raters from the Climate Fresk community, an NGO organizing workshops to explain IPCC reports to the public.\footnote{\url{https://climatefresk.org/world/}.}. We call this group the \emph{Advocates} (cf. \Cref{sect:app:cf}).

\subsection{Questions}

\name's questions are sampled from the logs of \href{www.chatclimate.ai}{\texttt{chatclimate.ai}}~\citep{vaghefi2023chatclimate} a chatbot dedicated to climate change which has collected thousands of users' questions. 
These questions are challenging because they involve technical topics with complex interdisciplinary ramifications. Compared to trivia and puzzles, real users' questions express genuine information needs. At the same time questions can be poorly worded or unclear, making it difficult to even interpret the question or the context in which it is posed. Furthermore, the role of evidence is critical. The ability to properly assess answers to such questions may be directly useful for advancing AI-assisted science, science communication and decision-making in the real world.

To match our Scientist group’s expertise, we selected questions from six key topics: ’Weather and Climate Extremes’, ’Mitigation Pathways’, ’Detection, Attribution and Uncertainty’, ’Climate Finance and Risks’, ’Climate Impacts, Adaptation and Vulnerability’, and ’Climate Change Scenarios’.
For each topic, Scientists chose approximately 30 questions and assigned one of three difficulty labels: High Confidence (answerable from authoritative sources), Advanced (requiring technical sophistication), or Open (actively debated). We also classified each question by its relevant IPCC working group (WGI: Science, WGII: Adaptation, WGIII: Mitigation).\footnote{Cf., \url{https://www.ipcc.ch/working-groups/}.}
The final dataset represents all topics and working groups, with a slight prevalence of climate science questions (WGI) and the ’Impacts’, ’Detection’, and ’Scenarios’ topics. High Confidence ($35.7\%$) and Advanced ($35.1\%$) questions are the most frequent, though the fraction of Open questions is substantial ($29.2\%$). These properties are summarized in \Cref{fig:question-topic-wg-level} (Appendix).

\subsubsection{Answer Format}
Answers include three components: \emph{text}, \emph{images} (optional), and \emph{references}.  
Typically, 500 words are sufficient for an initial, yet substantial, answer and we use this as the length limit for the main body of the response.\footnote{Answers should be concise and to the point, also to lighten the cognitive load on human raters. A 500 words answer can typically be displayed in full on most screens without scrolling.}
All key points in the answer that quote, explicitly or implicitly, external sources must be accompanied by citations, with the references listed in a dedicated section. We encourage the inclusion of images, e.g.\ to summarize quantitative information.

\subsection{Data Construction Process}
As in recent work \citep{ruan2025expertlongbench,arora2025healthbench}, we design a rubric-based assessment task. Rubrics provide a level of abstraction that can contribute to scalable, accurate and consistent, model-based evaluation, moving beyond "vibe-based" assessments \citep{robertson2025let}. We want the rubrics to be data-driven, specific to technical points, common pitfalls and misconceptions.  To this end, we implement a data processing pipeline, see \Cref{fig:placeholder}, which emphasizes a transparent and traceable curation and assessment process. 

\begin{figure*}[t]
    \centering
    \begin{subfigure}[b]{0.46\textwidth}
        \centering
        \includegraphics[width=\textwidth]{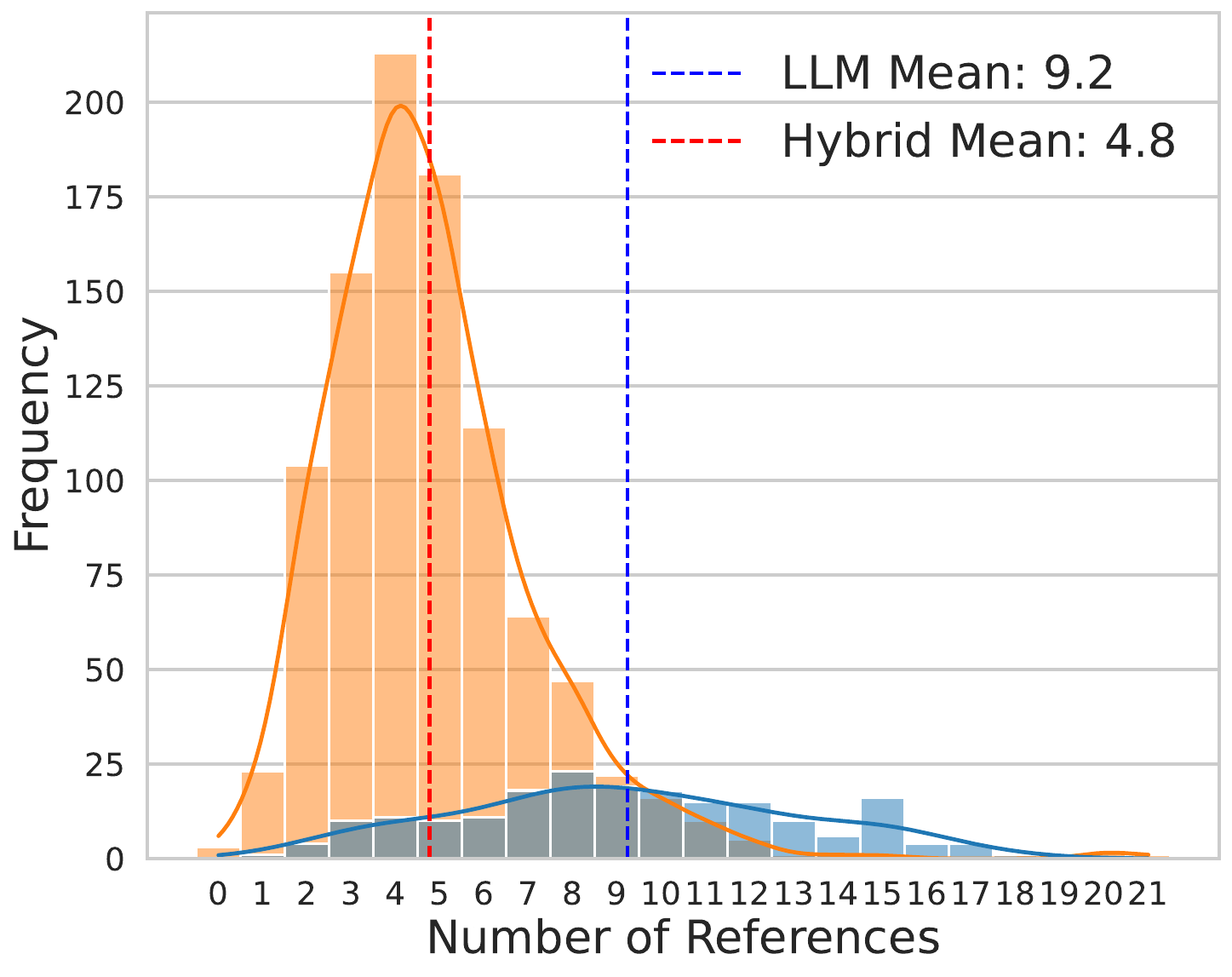}
        \subcaption{References counts in candidate answers.}
        \label{fig:subgraph1}
    \end{subfigure}
    \hfill
    \begin{subfigure}[b]{0.53\textwidth}
        \centering
        \includegraphics[width=\textwidth]{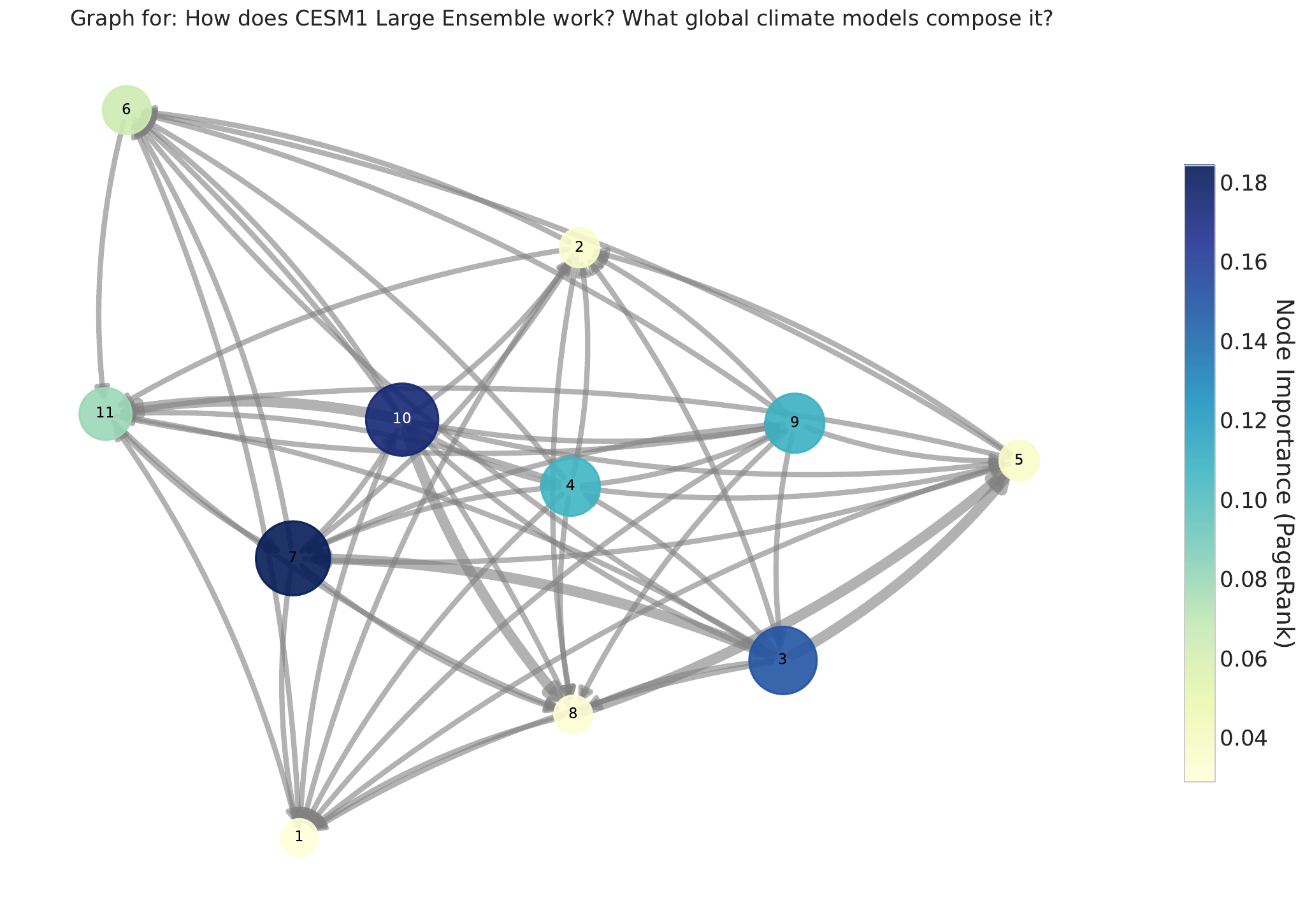}
        \subcaption{Graph with weighted edges and ranked nodes.}
        \label{fig:subgraph2}
    \end{subfigure}
    \caption{Left: Hybrid answers have more references than LLM answers. Right: Example of a pairwise preference graph for a single question.}
\end{figure*}

\paragraph{Phase 1 - Candidate answers:} For each question, a human produces a candidate answer in two steps: (i) by curating an \emph{outline} (ii) by curating a full \emph{first draft} answer derived from the outline. The first versions, of both outline and full draft, are generated by the model following an iterative self-improving, retrieval-augmented, process. 
We produce $N{\geq}3$ hybrid answers for each question. $92\%$ of the candidates are curated by Experts, $8\%$ by Advocates.
We add to the set an \emph{LLM answer}, produced independently by the model, and 
a \emph{merged} answer. This is produced by the model by synthesizing the existing candidates. The total is $N{+}2$ answers. Overall, we collect 1330 candidate answers. $82.6\%$ have images. $99.7\%$ have references -- 6.1 on average; LLM answers have roughly twice as many references as hybrid answers (\Cref{fig:subgraph1}). See also \Cref{sect:data-candidate-answers} for more details. 

\paragraph{Phase 2 - Pairwise Answer Preferences}
\label{sect:phase-2}
To learn the different candidate answers' strengths and weaknesses, we run side-by-side evaluations where humans evaluate which answer in a pair is better and why, a standard approach for collecting human preference data \citep{chiang2024chatbotarenaopenplatform}. We provide a form, derived from~\citep{assessinglargelanguagemodels} where the graders can select aspects where either answer is superior. At least one dimension must be selected on the preferred answer side. 
Humans do not evaluate answers they have curated. We collect 8654 preferences, 2.17 on average per unique answer pair. $70\%$ of the pairwise preferences are from the Experts and $30\%$ from the Advocates.

Due to the scarcity of human raters, it is not possible to collect numerous preferences on answer pairs. In addition, due to the complexity and nuances of the task, we do not expect that this process would converge to the identification of clear-cut good and bad answers.
Instead, we deal with the topic of the question holistically, relying on the full set of answers and the intrinsically multi-dimensional human preference feedback.
We represent the pairwise preferences as a directed graph with edges weighted by the preference counts. This provides tools for computing characteristics of the data that can be used to gain insights for compiling the grading rubrics. \Cref{fig:subgraph2} plots an example showing how the graph structure can be complex and densely connected. 
Most graphs form one large strongly connected component, indicating the presence of intransitivity or low confidence regions. 
However, one can still find informative structure, such as sinks and sources (worse and better answers) and groups of high/low ranking answers.

\paragraph{Phase 3 - Question Rubrics}
In the last phase of the process, each question is associated with a \emph{rubric}, to be used for assessing answers in a transparent and consistent way. Our rubrics consist of a 'Cheat Sheet': a compact reference for what to look for in the answers, and a 'Grading Rubric', explaining the grade bands. We find the use of explicit grading useful, even if for practical reasons we mostly rely on pairwise evaluations, because it forces the explanation of decisions in a more interpretable and calibrated way. 
An initial draft of the rubric is generated by Gemini 2.5 Pro, using a prompt which lists all the answers, including images and references, the quality guidelines, the pairwise preference graph and other instructions.

\Cref{fig:subgraph2} shows the preference graph for the question "How does CESM1 Large Ensemble work? What global climate models compose it?". The graph forms one strongly connected component, indicating disagreements, mostly deriving from different prioritization of criteria among human raters. 
However, the model is capable of surfacing critical feedback from raters. For example, the model identifies the possible confusion between correct and incorrect atmospheric model components. To break ties and propagate non-local preferences, we rank answers/nodes. The model is able to propose an holistic synthesis of the data in the generated rubrics, including using the images content (cf. the example in~\Cref{sect-app:rubrics}).

The last step involves the Scientists, who manually curate the final form of the rubrics.
In general, the AI-generated rubrics provide an impressive starting point. Areas for improvement concern greater scientific nuance and rigor: sometimes the rubrics lacked depth or overlooked critical details -- e.g, including overshoots scenarios for a question on limiting warming to below a certain target. Initial rubrics can also reflect an overconfidence bias present in the published literature, where positive findings are over-represented. Another aspect to improve upon is the critical assessment of the question's premises, whether the question makes sense, is well-formulated and clear.

\subsection{AI-Assisted Curation Tool}
\label{sec:editor}
All steps above are performed in a dedicated AI-assisted user interface, called the \emph{\editor{}}. The purpose is to both facilitate and analyze the 'human-in-the-loop' process. The tool allows experts to carry out all the three phases in dedicated workflows supported by a context-aware LLM assistant. Additionally, the \editor{} provides direct access to web search for both documents and images, as well as ranking and understanding (e.g., summarization) of the retrieved content. For publications listed in OpenAlex\footnote{\url{https://openalex.org/}.}, we include bibliographic metadata. The tool accepts feedback in both structured and natural language. We use Gemini 2.5 Flash as the assistant AI 
and Google for web search. 

%% file: sections/4-results.tex
We evaluate several publicly available models: OpenAI's GPT-5 and o3, Google's Gemini 2.5 Pro and Flash, and Anthropic's Claude Opus 4.1 and Sonnet 4. Models are accessed via their public APIs with default settings, without search enabled. For each system, we submit one request per question in the dataset using a 'system prompt'. We also evaluate the highest ranking hybrid answer from our dataset. The 'system prompt' provides the question, information about the expected answer format, the quality dimensions, the role of references, visual content etc. The prompt explicitly states that each piece of evidence must be accompanied by a URL (cf. the full prompt text in \Cref{sect:sys-prompt}). 

\subsection{Model-Based Pairwise Assessment}
\label{sect:autorater}
We perform model assessment of answer pairs: two answers are evaluated side by side (SxS) to identify the preferred one. We adapt the Chatbot Arena's battles setup~\citep{chatbotareana}. A single battle involves one question and two systems answers. We run three rounds of evaluations. In each round, we execute two pairwise evaluations, swapping the order of the answers  to control for position bias \citep{wang2023large}. The system with the majority score is the winner; otherwise, it is a tie. This procedure defines a single battle. We compute battles between all pairs of sources answers, 4147 battles total. We estimate the ELO scores for all systems using the Bradley-Terry model, including a bootstrap $95\%$ confidence interval.

As the judge model we use \geminipro with a temperature of $0.7$. The model relies on an 'evaluation prompt' which includes the general quality guidelines, the question-specific rubric and cheat sheet, general rubrics, the input data (question, answer pair) and the task instructions. The latter include a required explanation for the decision aligned with the quality dimensions of~\Cref{sect:phase-2}. Images are encoded in place, in the answer, as bytes, if the image content can be fetched via the provided link. To assess the evidence we check the validity of the images and references links in the responses. We classify the returned status as either a valid web page, whose content may or may not be accessible; e.g., due to paywalled content, or as invalid, due to the URL being hallucinated. 
We instruct the model to factor in this information while assessing the answers’ statements. See \Cref{sect:eval-prompt} for the full prompt. We refer to the judge model as the \textbf{\name autorater}.

\subsection{Experiments Findings}\label{sec:expfindings}

\begin{table*}[t]
    \centering 
   \footnotesize
    \begin{tabular}{lrrrrrr}
    \toprule
          & \multicolumn{6}{c}{\name Autorater}\\
          \cmidrule(lr){3-7}
   & &  Answer SxS &\multicolumn{4}{c}{Answer Dimensions} \\
         \cmidrule(lr){3-3} \cmidrule(lr){4-7} %
        System &  &  &  Citations & Images & Knowledge & Presentation\\
        \midrule
        GPT-5           && \rankone 1150 $\pm$ 19 & \ranktwo 1104$\pm$9 & \rankfour 905 $\pm$ 33 & \rankone 1167 $\pm$ 8 & \ranktwo 1106 $\pm$ 15 \\
        Claude Opus 4.1  && \ranktwo 1135 $\pm$ 19 & \rankone 1219$\pm$10 & \rankthree 965 $\pm$ 22 & \ranktwo 1153 $\pm$ 7 & \rankfour 954 $\pm$ 13 \\
        GPT o3          && \rankthree1018 $\pm$ 18& \rankseven 846 $\pm$ 7 & \rankseven 785 $\pm$ 33 & \rankthree 1066 $\pm$ 7 & \rankone 1349 $\pm$ 23 \\
        Gemini 2.5 Pro  && \rankfour 969 $\pm$ 18 & \rankfour 949 $\pm$ 8 & \ranktwo 970 $\pm$ 19 & \rankfour 954 $\pm$ 6 & \rankthree 960 $\pm$ 20 \\
        Hybrid          && \rankfive 945 $\pm$ 18 & \rankfive 913 $\pm$ 7 & \rankone 1358 $\pm$ 21 & \ranksix 868 $\pm$ 6 & \rankseven 749 $\pm$ 16 \\
        Claude Sonnet 4 && \ranksix 915 $\pm$ 19  & \rankthree 981 $\pm$ 7 & \rankfive 822 $\pm$ 30 & \rankfive 885 $\pm$ 6 & \rankfive 861 $\pm$ 19 \\
        Gemini 2.5 Flash&& \rankseven 868 $\pm$ 18& \ranksix 875 $\pm$ 8 & \ranksix 798 $\pm$ 30 & \rankseven 813 $\pm$ 7 & \ranksix 803 $\pm$ 15 \\
       \midrule
             & \multicolumn{6}{c}{Human Assessment}\\
          \cmidrule(lr){2-7}
          & \multicolumn{2}{c}{Answer SxS} &\multicolumn{4}{c}{Answer Dimensions (Experts)} \\
         \cmidrule(lr){2-3} \cmidrule(lr){4-7} %
        System & Experts & Scientists &  Citations & Images & Knowledge & Presentation\\
        \midrule
         GPT-5         & \rankfive$906\pm18$ & \ranktwo$1040\pm115$ & \rankthree$970\pm8$ & \rankfive$633\pm31$ & \rankthree$975\pm8$ & \rankfive$902\pm8$ \\
         Claude Opus 4.1      & \rankone$1115\pm20$ & \rankone$1149\pm111$ & \rankone$1244\pm12$ & \rankthree$862\pm22$ & \ranktwo$1067\pm8$ & \rankone$1098\pm9$\\
         GPT o3        & \rankfour$950\pm20$ & \rankthree$959\pm114$ & \rankfive$808\pm9$ & \rankfour $716\pm27$ & \rankone$1078\pm8$ & \rankthree$935\pm7$ \\
          Gemini 2.5 Pro & \ranktwo$1015\pm18$ & \rankfour$943\pm144$ & \ranktwo$1043\pm9$ & \ranktwo$1042\pm20$ & \rankfour$970\pm7$ & \ranktwo$1062\pm8$ \\
        Hybrid    & \rankthree$1015\pm20$ & \rankfive$910\pm115$ & \rankfour$842\pm8$ & \rankone $1486\pm25$ & \rankfive$831\pm9$ & \rankfour$922\pm8$\\
         \bottomrule
    \end{tabular}
    \caption{ELO scores and 95$\%$ confidence intervals for the \name Autorater and human assessments. The models in the lower table are sorted in the same order as in the upper table.}
    \label{tab:results-main}
\end{table*}

\Cref{tab:results-main} summarizes the experimental results.\footnote{In \Cref{app:examples}, we discuss specific examples of question-answer pairs and their evaluation.} We report ELO scores for overall pairwise preference (Answer SxS) and per quality category dimension (Answer Dimension). The upper table contains the results for the \name autorater. The lower table reports the results from a manual validation experiment for a sample of 1976 battles from the top 5 systems outputs, using the protocol of Phase 2 (\Cref{sect:phase-2}). This manual assessment, 'Answer SxS/Experts' and 'Answer Dimensions (Experts)' in \Cref{tab:results-main}, relies only on the Experts group, three raters per battle. Since the autorater and the Experts reveal systematic disagreements we sampled a number of disagreement battles for deeper analysis. The Scientists repeated the pairwise human evaluation for 72 of these battles, see 'Answer SxS/Scientists' in \Cref{tab:results-main}.\footnote{\Cref{sect:app:detailed-dimensions} and \Cref{sect:app:question-type} report results at more granular quality dimensions and by question types, which broadly align with the general findings reported here.}

\paragraph{Humans/Autorater Agreement}
The main disagreements between Experts and the \name Autorater concern the quality of the \hybrid{} and \geminipro{} answers, on one side, and that of \gpt{}, on the other. On close inspection, we find the Scientists to agree with the \name autorater. In particular, they prefer \gpt{} and \opus{} for 'Knowledge' and 'Presentation'. %
We are inclined to attribute the Experts preference to familiarity bias, due to overexposure to Gemini and hybrid outputs in Phase 1 and 2 of the data creation. The Experts may have become accustomed to the presentation style of Gemini, while that of \gpt{} is more terse and relies more heavily on mathematical notation. 
\gpt{}, and also \othree{}, are also penalized by the Experts for the lack of images, a strength of the Hybrid and \geminipro{} answer, which is however a secondary epistemological factor for the Scientists.

\paragraph{Hybrid Answers} \hybrid{} answers are better than those of the underlying LLM, \geminiflash{} -- which is consistent with results from Phase 2 (\Cref{sect:data-candidate-answers}). They are also better than \sonnet{}'s answers, and not far behind \geminipro.
However, \hybrid answers are ranked lower than top models, by both autorater and Scientists. They are worse in terms of knowledge and presentation by autorater, Scientists and Experts. They rank lower also in presentation and citations quality. This indicates that the quality of the model in the human-in-the-loop framework is crucial, and better models exceed the quality of hybrid methods based on lower-performing models. 
We also note that the Advocates group produced, in Phase 1, high-quality answers. Experts rated the Advocates' answers higher than any other answer source, including their own (cf. \Cref{sect:app:cf-quality}). Evidence from the \editor{} use shows that Advocates engaged much more with all aspects of curation, including AI assistance. This suggests that high motivation and AI assistance can be an effective combination.

\paragraph{Frontier Models' PhD-Level Knowledge}
Based on the experiments, and close qualitative inspection, it is clear that the frontier models' knowledge quality is remarkably high. On the \name data, in terms of knowledge and presentation, \opus{}, the two OpenAI models and \geminipro{}, match or exceed the performance of PhD level humans. 

\paragraph{Ablations} We perform several ablation studies with the autorater (\Cref{tab:elo_controls_main}). Notably, removing the question-specific rubrics from the prompt changes the results only in the bottom half, with the \hybrid{} answers overtaken by \geminiflash{} and \sonnet{}. This suggests that the additional resolution provided by the rubrics applies primarily to the kind of responses used to develop the rubrics. Or, in other words, that rubrics are far from complete. Hence, it is important that rubrics adapt to new data as better models become available. 
While the system-level scores do not change much without rubrics, we notice a trend in the justifications of individual ratings. The autorater with rubric often assigns high scores to \emph{ungrounded} information that correctly matches the rubric requirements, while the rubric-free autorater follows the general requirement to treat any claims that are not supported by references as non-existent more strictly. Thus, removing the rubrics slightly shifts the assessment focus from 
knowledge quality towards mechanistic verifiability.

Other ablations show that the autorater moves the judge model away from intrinsic biases, including favoring style over substance \citep{gudibande2024false} and AI-AI bias \citep{panickssery2024llm}. The evidence instructions and the quality of the judge model have also a large influence.

\begin{figure*}[t]
    \centering
    \begin{subfigure}{0.95\textwidth}
        \centering
        \includegraphics[width=\linewidth]{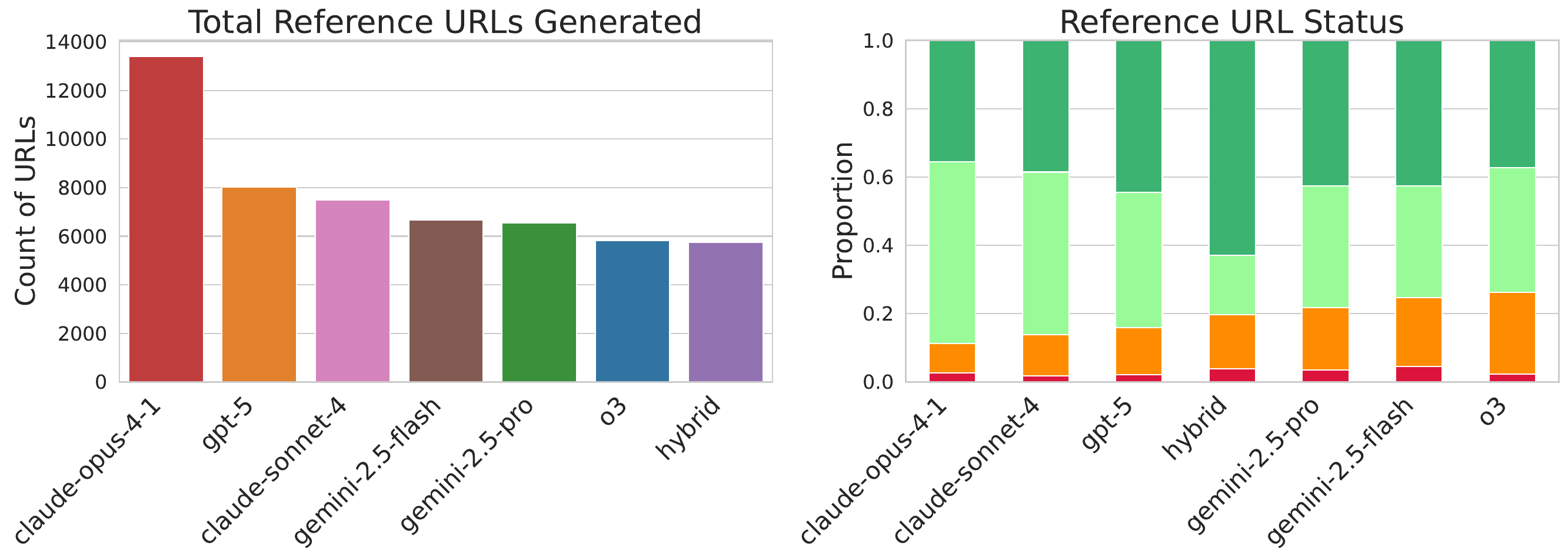}
    \end{subfigure}

    \vspace{0em} 
    
    \begin{subfigure}{0.95\textwidth}
        \centering
        \includegraphics[width=\linewidth]{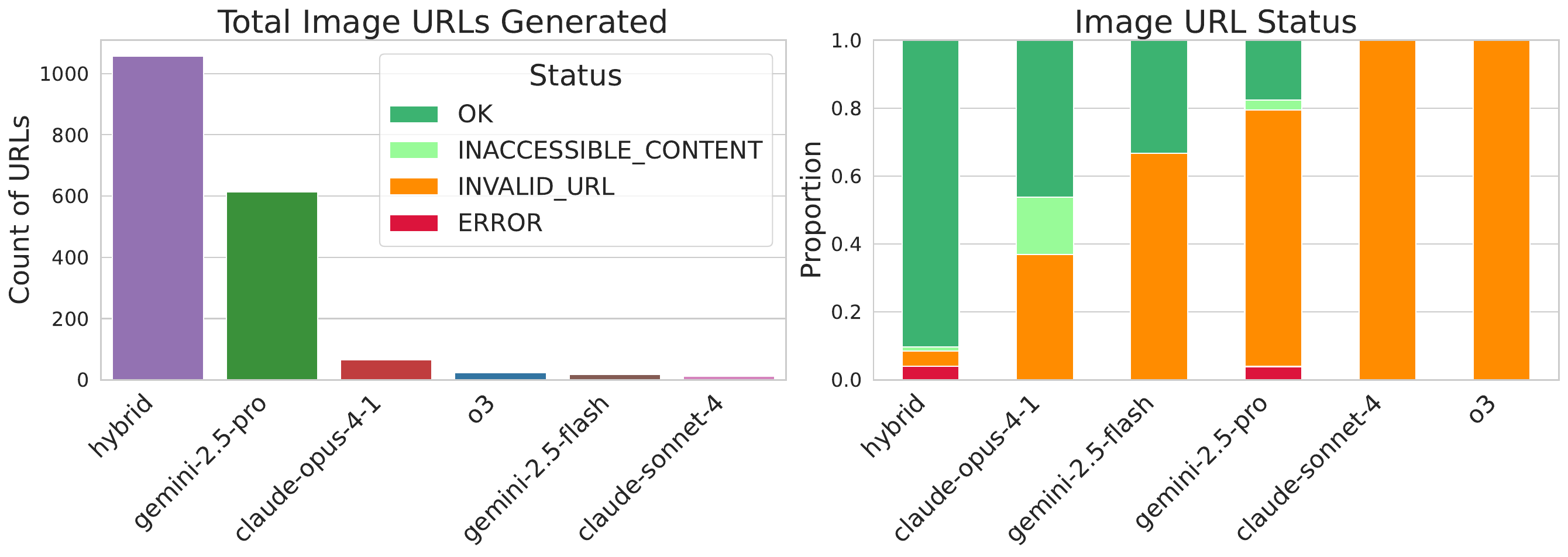}
    \end{subfigure}
    \caption{Number of reference (top), and image (bottom), URLs and their status.}
    \label{fig:evidence-status}
\end{figure*}

\paragraph{Evidence Quality}
\Cref{fig:evidence-status} summarizes numbers and status of evidence links. \opus{} is by far the best for quantity and validity of citations. It has also the highest rate of pay-walled content, likely peer-reviewed sources. It is followed by \gpt{}, \sonnet{} and \geminipro{}. Roughly $25\%$ of \othree's URLS are hallucinated, corroborating recent findings on the persistence of citation errors in LLMs \citep{byun2024reference, tang2024citeeval}. Experts and the \name Autorater give low scores to \hybrid answers (which have the fewest citations), highlighting the gap between AI and humans in terms of literature processing capacity. 

For images the results are even clearer. Hybrid answers stand out for superior visuals, while LLMs' answers often have very poor choice of visuals, or none. Multimodal understanding and composition shows considerable headroom for models. Among models, Gemini 2.5 Pro has the best performance on image quality according to both Experts and autorater. It generates a  number of hallucinated URLs but also far more links than the next system, \othree{}. The \name autorater image scores for \gpt{} seem somewhat unjustified, given that it does not provide image links and may point to \emph{anchoring bias}: the judge model has a hard time penalizing an overall strong output. If the inclusion of images is made mandatory, the hallucinated links rates range between $50\%$ and $80\%$ (\Cref{fig:image_specific_url_status_control7}).

\paragraph{Informal assessment of strengths and weaknesses}
In addition to pair-wise preferences the autorater also produces a detailed assessment across question specific and shared rubric dimensions which can be used for further analysis. 
As the amount of text is too large for manual analysis, we use \geminipro to extract recurring patterns and examples from the justification.
Hence the characterizations in Table \ref{tab:model_comparison} and Appendix \ref{app:characters}  should be considered \emph{qualitative and informal}, however they are in line with the results of \Cref{tab:results-main} and \Cref{sect:app:detailed-dimensions}.

\begin{table}[htb]
\centering
\footnotesize
\caption{Informal strengths (+) and weaknesses (-). See also Appendix \ref{app:characters}.}
\label{tab:model_comparison}
\begin{tabular}{@{}lccccc@{}}
\toprule
 & \gpt & \othree & \opus & \geminipro & \hybrid \\ \midrule
Question Interpretation & & & - & + & + \\
Facts and Numbers & + & + & + & - & -  \\
Depth \& Nuance & + & + & + & - & - \\
Grounding \& Images & & - & + & + & +  \\
\bottomrule
\end{tabular}
\end{table}

%% file: sections/5-related-work.tex
As foundation models achieve high performance on general knowledge tests like MMLU~\citep{hendrycks2020measuring}, the research community has pivoted to more challenging evaluations to differentiate model capabilities. This includes benchmarks testing advanced reasoning with graduate-level or ``Google-proof'' questions, such as GPQA~\citep{gpqa}, HLE~\citep{hle}, and OlympiadBench~\citep{he2024olympiadbench}. Concurrently, there is surging interest in "AI for Science"~\citep{eger2025transforming}, leading to the development of specialized "Science LLMs"~\citep{Zhang2024ScientificLLA} and benchmarks across diverse domains. This includes efforts in medicine~\citep{arora2025healthbench}, biology~\citep{laurent2024lab, justen2025llmsoutperformexpertschallenging}, chemistry~\citep{Bran2023AugmentingLLA, Malikussaid2025BridgingTPD}, and the development of AI research assistants or "AI Scientists"~\citep{Xie2025HowFAC, Liu2025TowardsAIE}. This trend extends to specific scientific skills, such as multimodal understanding and figure interpretation (MMMU~\citep{mmmu}, SciFIBench~\citep{roberts2024scifibench}), scientific coding (SciCode)~\citep{tian2024scicode}, and tool-augmented scientific reasoning (SciAgent)~\citep{ma-etal-2024-sciagent}. While these benchmarks establish a high level of difficulty, often claiming expert-level performance, they frequently rely on multiple-choice or short-answer formats for algorithmic verifiability~\citep{justen2025llmsoutperformexpertschallenging, Dinh2024SciExBLA}. In the domain of climate change, there is work on AI alignment~\citep{kaack2022aligning, rolnick2022tackling}, and efforts like ClimaQA~\citep{manivannan2024climaqa} and AtmosSci-Bench~\citep{li2025atmosscibenchevaluatingrecentadvance} that use expert-level content but are similarly constrained by traditional formats. \name fills a critical gap by focusing on the complexity of synthesizing and communicating established scientific knowledge~\citep{Bajpai2024CanLRA} in response to real user queries~\citep{vaghefi2023chatclimate}.

A central challenge for generative models is producing extended, coherent, and verifiable text. Early Long-Form Question Answering (LFQA) datasets like ELI5~\citep{fan2019eli5} established this task, often addressed using Retrieval-Augmented Generation (RAG) frameworks~\citep{lewis2020retrieval}. This evolved into models explicitly designed to generate answers supported by retrieved evidence and citations, such as WebGPT~\citep{nakano2021webgpt} and GopherCite~\citep{menick2022teaching}. However, ensuring faithfulness and avoiding hallucination remain core problems~\citep{ji2023survey}. The need for reliable grounding has led to a proliferation of research focused on citation and attribution. Recent studies highlight persistent issues, including the generation of non-existent references~\citep{byun2024reference} and poor citation quality, particularly in long-context scenarios~\citep{tang2024citeeval, zhang-etal-2025-longcite}. This has motivated new benchmarks, including general-purpose evaluations like ALCE~\citep{gao2023enabling} and ASQA~\citep{stelmakh2022asqa}, as well as specialized frameworks focusing on long-context QA (LongCite~\citep{zhang-etal-2025-longcite}, L-CiteEval~\citep{tang2024citeeval}) and medicine (MedCite~\citep{wang-etal-2025-medcite}, \citep{wu2025automated}). Various methods aim to improve attributed generation, such as Chain-of-Thought prompting~\citep{ji2024chain}, preference learning~\citep{DBLP:conf/acl/LiSHLH0Z24}, self-reflection and critique~\citep{asai2024selfrag}, and transparent utilization of internal and external knowledge~\citep{shen-etal-2025-transparentize}. Beyond autonomous generation, \name also investigates answers created through Human-AI collaboration, a critical area of study as AI integrates into knowledge work~\citep{treude2025developers}. Research shows that while AI assistance can improve productivity, the quality of the base model and the nature of the collaboration significantly impact the final output~\citep{noy2023experimental}. Furthermore, the methodology of using experts-in-the-loop to create the benchmark aligns with dynamic benchmarking strategies designed to keep pace with model advancements~\citep{kiela2021dynabench}. While benchmarks like ExpertLongBench~\citep{ruan2025expertlongbench} and HealthBench~\citep{arora2025healthbench} also use rubric-based evaluation, \name is unique in its focus on a combination of synthesis, attribution, multimodality, and hybrid generation.

Evaluating complex outputs at scale has motivated the "LLM-as-a-Judge" paradigm as a scalable complement to human evaluation~\citep{zeng2023evaluating, li2024llms, gu2024survey}. 
Strong models can achieve high alignment with human preferences in pairwise comparisons~\citep{zheng2023judging}, leading to methodologies like comparative assessments with ELO rating systems, popularized by Chatbot Arena~\citep{chatbotareana}. However, the reliability of this paradigm is a subject of intense research, with efforts dedicated to benchmarking the judges themselves for ranking capabilities~\citep{gera2025justrank}. Studies have identified critical inherent biases, including position bias (favoring the first-presented answer), verbosity bias, and self-enhancement bias~\citep{wang2023large, wang-etal-2024-large-language-models-fair}. Furthermore, recent work has shown a systematic ``AI-AI bias,'' where models prefer LLM-generated text over human-authored text, regardless of objective quality~\citep{laurito2025aiai, panickssery2024llm}, and that pairwise comparisons can amplify these biased preferences~\citep{jeong2024comparative}. Other work cautions that high alignment may be misleading, as judges might favor stylistic similarity over substantive correctness~\citep{gudibande2024false}, that critiques generated by judges can be flawed~\citep{sun-etal-2024-critique}, and that a model's ability on a task does not guarantee its ability to evaluate it~\citep{oh-etal-2024-generative}. This growing body of evidence highlights the fragility of naive LLM-as-a-Judge implementations, and the need for rigorous safeguards.

In response to these challenges, our work aligns with a research thrust focused on increasing the methodological rigor of LLM-based evaluation~\citep{Calderon2025TheAAA}. The core of \name's contribution is a transparent, expert-in-the-loop process for generating fine-grained, per-question rubrics, moving beyond "vibe-based" assessments~\citep{robertson2025let}.
This approach can be viewed as a form of scalable oversight, leveraging model assistance to apply expert-defined criteria consistently~\citep{bowman2022measuring}. This criteria-based approach is validated by several studies showing improved alignment with human experts. \citet{kim2024prometheus} (Prometheus) trained a specialized evaluator LM on custom rubrics. Other approaches focus on decomposing evaluation into atomic facts for fine-grained assessment (e.g., FActScore)~\citep{min2023factscore}. Other frameworks leverage checklists (RocketEval~\citep{weirocketeval}) or combine rubric-based assessments~\citep{hashemi-etal-2024-llm} with reinforcement learning to mitigate biases, particularly in scientific domains (YESciEval~\citep{dsouza-etal-2025-yescieval}). Fine-tuning judges can also yield scalable and accurate evaluators (e.g., JudgeLM)~\citep{zhujudgelm}. The reliability of the judges themselves remains crucial, emphasized by benchmarks like JudgeBench~\citep{tan2025judgebench}. \name's methodology directly addresses the known biases of LLM judges by anchoring our evaluation against the highest level accessible of human domain expertise.

%% file: sections/conclusion.tex
Human and AI-based assessments suggest that frontier models have reached PhD-level performance in knowledge and presentation on advanced climate change topics. However, this competence is undermined by hallucinations of images and references. To improve traceability and trust, models must provide better explicit evidential support for their responses. 
Our next steps include evaluating search-enabled systems, making rubrics adaptive, and applying deeper scrutiny to evidence. As we envision the best future models to incorporate advanced skills, including dataset retrieval, statistical analysis, and image synthesis, the benchmarks have to co-evolve.

The evaluation of generative responses in deep-expertise regimes is a difficult, often ill-defined (Who is the audience? What was the original intent?), expensive, and fundamentally unscalable task for humans. We propose that collaborative frameworks, where experts and AI perform meaningful end-to-end tasks together, are key to scalable performance tracking. Climate change is a particularly well-suited domain for this approach, as its institutional practices are built on collaboration, consensus-building, and the rigorous assessment of evidence.

Our findings, however, reveal a critical challenge: while human-in-the-loop curation improves weaker models, frontier models can already surpass these hybrid results autonomously. A central challenge is the design of interfaces that enable expert-AI collaboration, perhaps through continuous interaction and mutual questioning, to achieve a synergistic performance that exceeds what the model can do alone. 
Such a framework inherently cultivates essential human skills: the critical assessment of AI-generated content, the verification of claims against ground-truth data, and the identification of model biases. Ultimately, this new mode of collaboration trains humans to be more sophisticated thinkers, strengthening our collective ability to validate information and build robust knowledge in an age of AI.

%% file: sections/20-appendix.tex
\section*{Appendix}

\input{sections/21-appendix-data}

\input{sections/22-appendix-human-experts}

\section{Results}
\label{sect:app-results}

\input{sections/24-appendix-ablations}

\input{sections/25-appendix-visuals}

\input{sections/26-appendix-question-type}

\input{sections/27-appendix-details-eval}

\input{sections/28-appendix-prompts}

\input{sections/29-appendix-characters}

\input{sections/30-appendix-examples}

%% file: sections/21-appendix-data.tex
\section{Data}
\subsection{Questions}
\label{sect:data-questions}
\begin{figure*}
    \centering
    \includegraphics[width=0.95\textwidth]{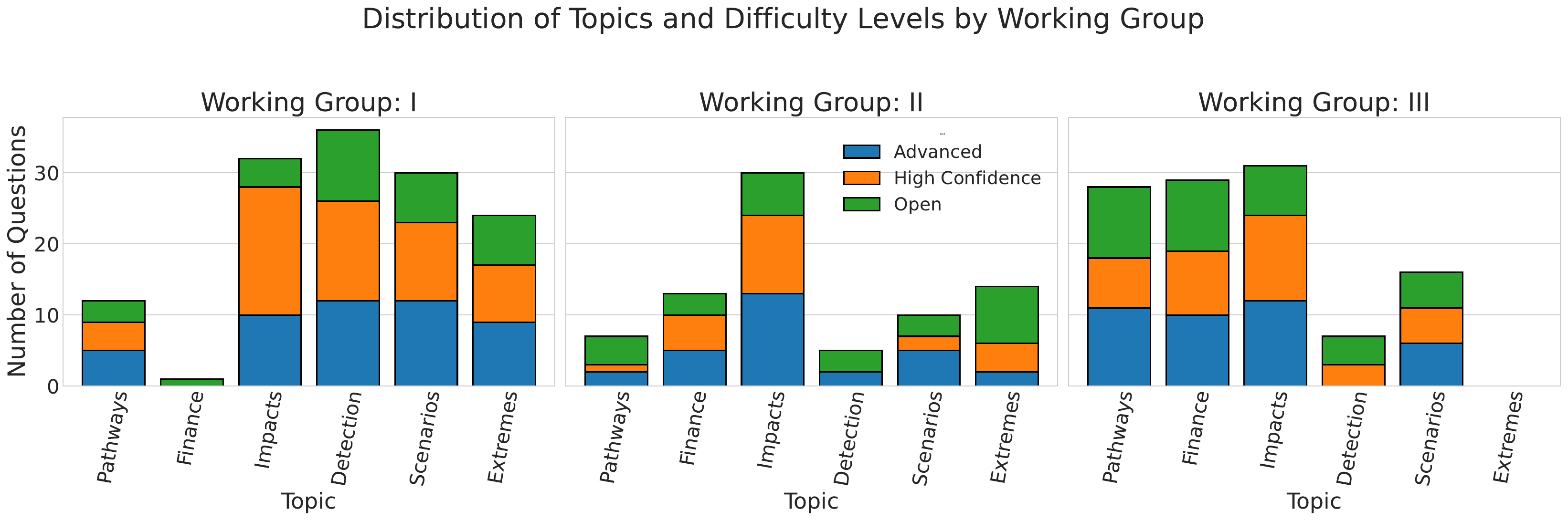}
    \caption{Distribution of topics and difficulty level by working group.}
    \label{fig:question-topic-wg-level}
\end{figure*}

The initial dataset consists of real-world user questions collected from \url{www.ChatClimate.ai}, a platform that provides climate-relevant information grounded in the IPCC Sixth Assessment Report (AR6). The questions represent authentic user inquiries submitted through the platform's conversational interface, which employs a Retrieval-Augmented Generation approach \citep{vaghefi2023chatclimate}. The data is publicly accessible on \url{https://doi.org/10.5281/zenodo.13788802}.

\Cref{fig:question-topic-wg-level} summarizes the main dimensions of the final \name questions set: IPCC working group, difficulty level and topic.\footnote{Each question can be assigned multiple labels, for each category.} 

\subsection{Candidate Answers}
\label{sect:data-candidate-answers}

\begin{figure*}[htb]
    \centering
    \begin{subfigure}[b]{0.49\textwidth}
        \centering
        \includegraphics[width=\textwidth]{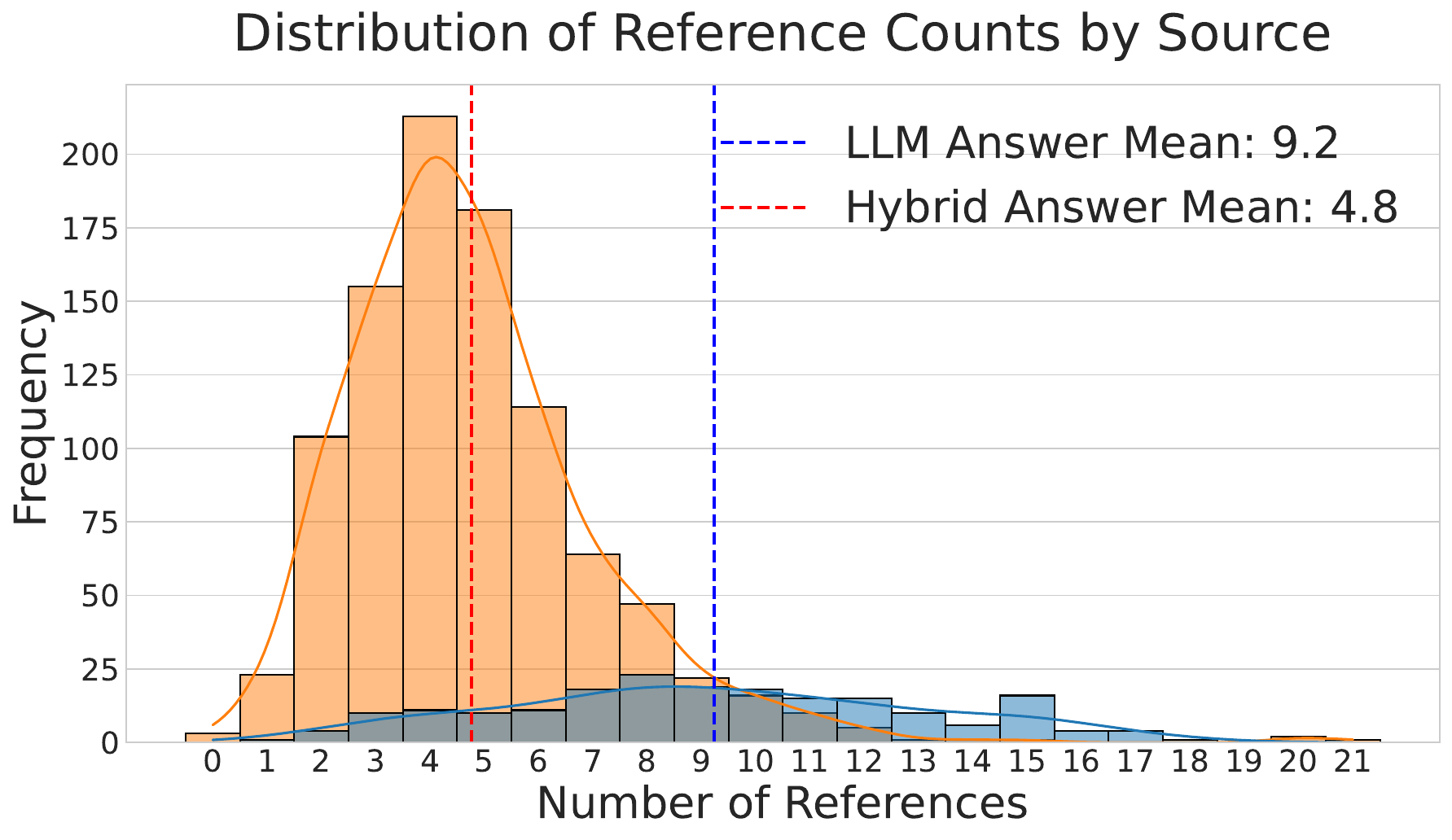}
        \subcaption{Answer references count distribution.}
        \label{fig:reference-hist}
    \end{subfigure}
    \hfill
    \begin{subfigure}[b]{0.49\textwidth}
        \centering
        \includegraphics[width=\textwidth]{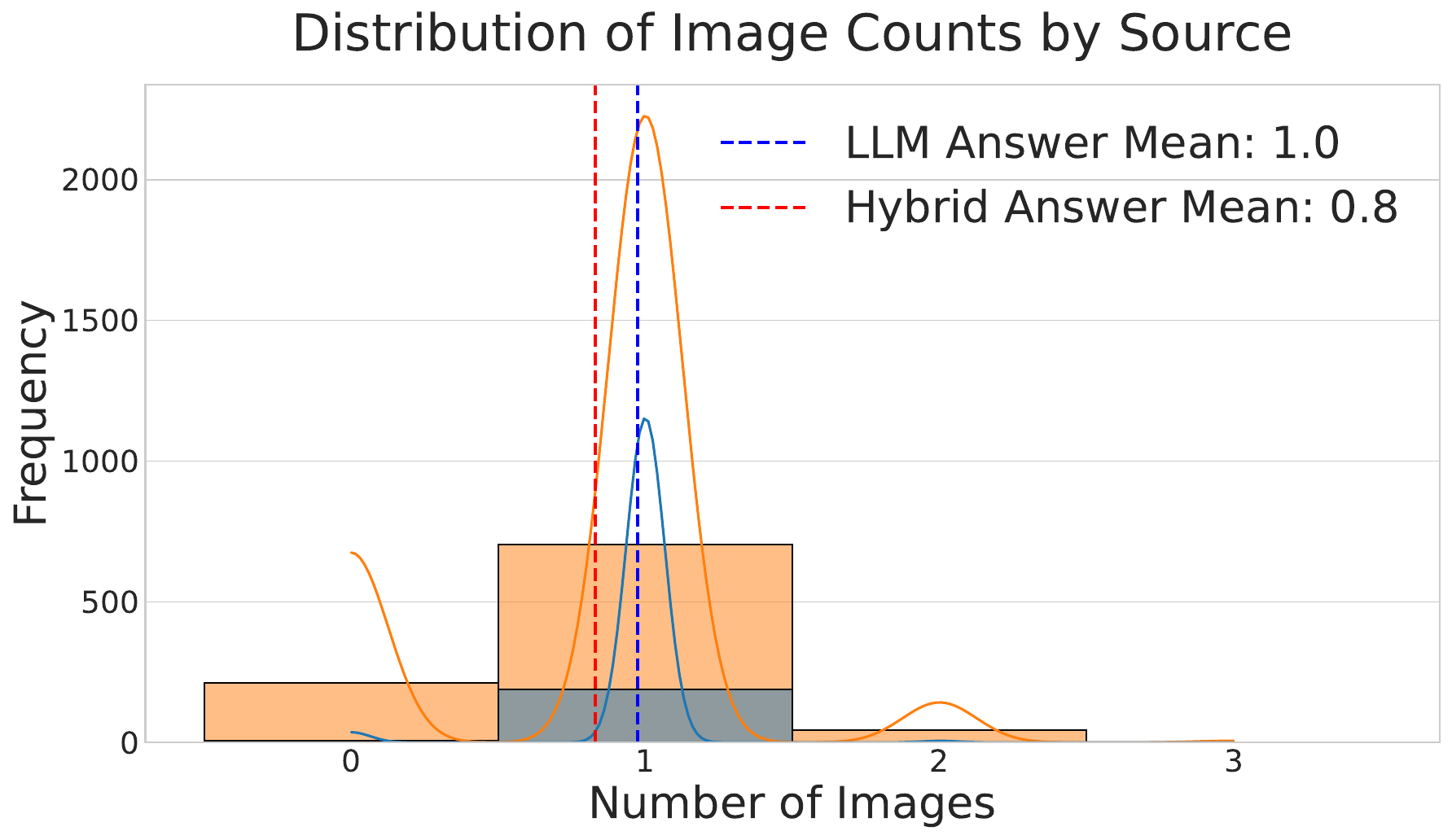}
        \subcaption{Answer image count distribution.}
        \label{fig:image-hist}
    \end{subfigure}
    
    \caption{Distributions of counts for references (a) and images (b) in hybrid and LLM answers.}
    \label{fig:reference-image-hists}
\end{figure*}
\Cref{fig:reference-image-hists} reports statistics relative to the number of references and images in the candidate answers data. Overall, there are 8086 references, the mean number of references for Hybrid answers is 4.8 and 9.2 for the 'LLM' answers, cf. \Cref{fig:reference-hist}. Hence, human experts tend to cut down substantially the references suggested by the LLM. The total number of images in the candidate answers set is 1176, the mean number of images in 'LLM' answers is 1.0 and 0.8 for the 'Hybrid' answers. \Cref{fig:image-hist} plots the frequency distributions. Also for images, humans remove images when they are not considered necessary, or a good image cannot be found. 

\subsubsection{LLM Answer Generation}
\label{sect:llm-answer}
\paragraph{Answer Outline}
The initial answer outline is created by prompting the model to generate a structured plan, limited to max 5 points, for writing a short essay on the topic of the question.
\paragraph{First Draft Answer}
A full answer is generated by prompting the model to transform the user-curated outline into a full-prose answer of approximately 300 words. The context regarding the interaction with the user while curating the outline, is included in the prompt.
\paragraph{LLM Answer}
The LLM answer added to the set of candidate answers is created as follows:
\begin{enumerate}
    \item An initial answer is generated using only the user's question as the prompt;
    \item We prompt the LLM to suggest search queries about the main keypoints in the answer;
    \item The queries are issued to Google, to obtain both document and image search results.
    \item The LLM filters out non relevant or low quality results and ranks the rest.
    \item The LLM improves the answer using, and citing as needed, documents and images.
\end{enumerate}

\paragraph{Merged Answer}
The 'Merged' answer is generated using the LLM, as follows:
\begin{enumerate}
    \item In the prompt we list the question and all the answers, including images and references;
    \item We ask the model to merge all the keypoints made in the answers in a coherent way.
    \begin{itemize}
        \item The model should resolve conflicting keypoints by consulting the supporting evidence.
        \item In the absence of a clear resolution, different views should be included and explained.
    \end{itemize}
    \item We ask the model to review, reconcile and renumber the references appropriately, so to keep only those required by the answer. Similarly for the image(s).
\end{enumerate}

\subsection{Pairwise Answer Preferences}
\label{sect:pairwise-answer-preferences}
\begin{figure*}
    \centering
    \includegraphics[width=0.95\linewidth]{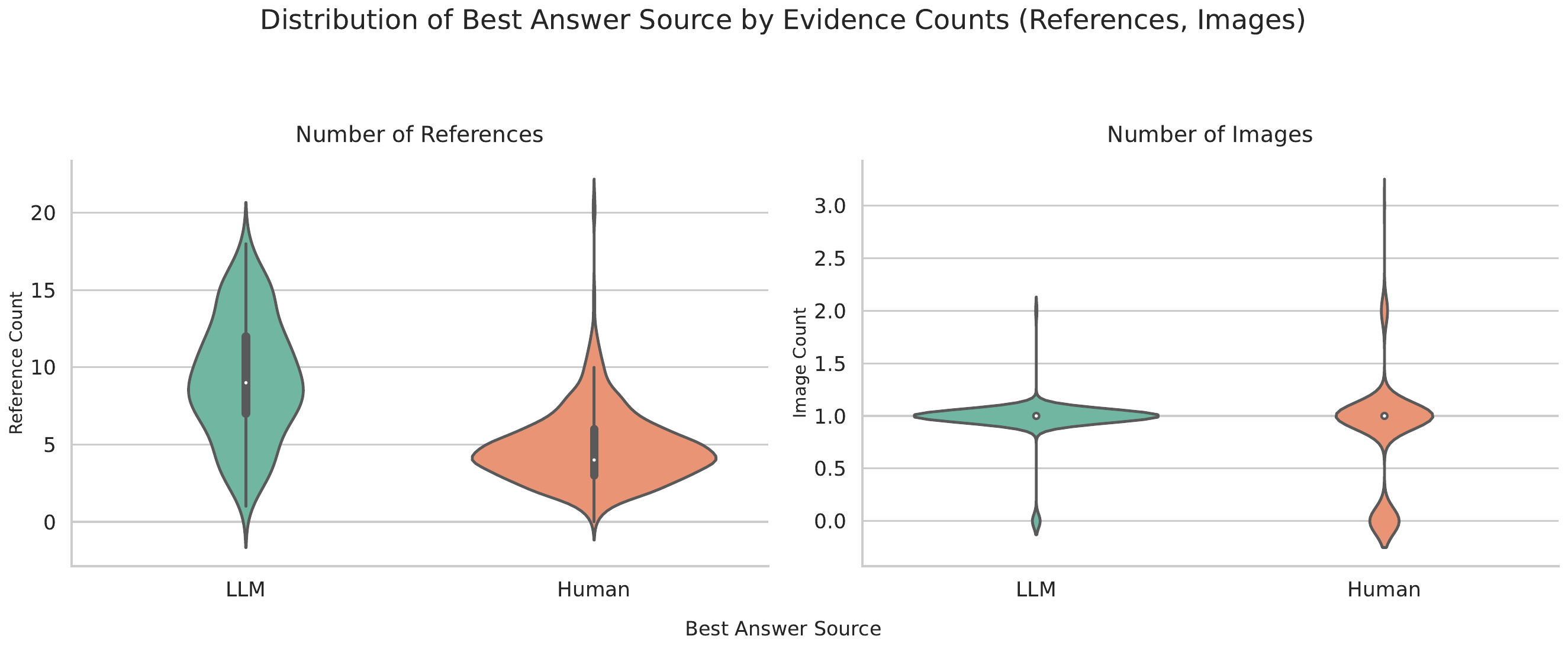}
    \caption{Distribution of best answer source, Hybrid or LLM, and evidence: images, references.}
    \label{fig:best-answer-by-evidence}
\end{figure*}

We evaluate the candidate answers quality from the preference graphs. The answer graph encodes preferences as directed edges between answer pairs, weighted by the counts on that preference. To rank the answers, we compute the pagerank score~\citep{Page1998PageRank} for each node. As ranking metrics, we report Mean Reciprocal Rank (MRR) and Precision@1. The table below shows that the Hybrid answer is the best 70.3$\%$ of the time, the LLM answer is preferred 19.1$\%$ of the time. The Merged answer is the least preferred (10.5$\%$). 
\begin{center}
    \begin{tabular}{c|cc}
    \toprule
        Answer Source & Precision@1 & MRR \\
        \midrule
        Merged & 10.5 & 0.32 \\
        LLM    & 19.1 & 0.42 \\
        Hybrid & 70.3 & 0.84 \\
    \bottomrule
    \end{tabular}
\end{center}

Best answers are not homogeneous in terms of evidence type and quantity, see \Cref{fig:best-answer-by-evidence}. As in the broader candidate answer distribution (\Cref{fig:reference-image-hists}), the best LLM answers have almost double the number of references as the Hybrid answers, on average. Similarly for images, removing an image doesn't necessarily make the answer worse. 

\begin{figure*}
    \centering
    \includegraphics[width=0.95\linewidth]{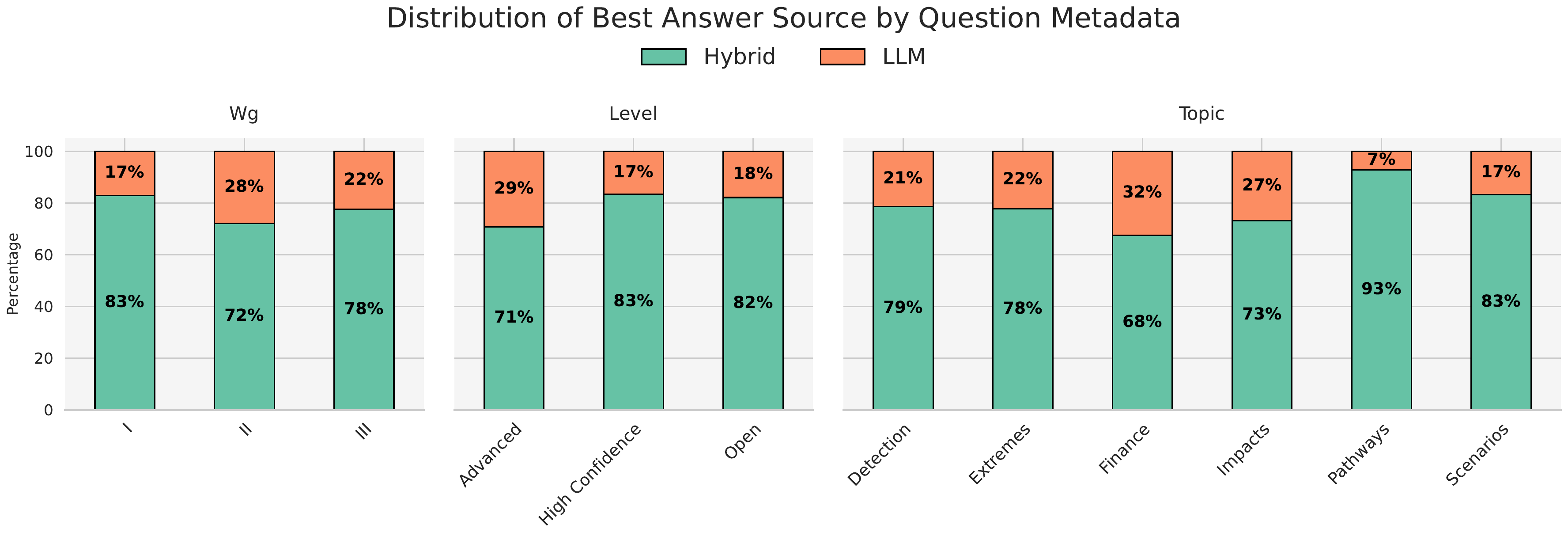}
    \caption{Distribution of best answer source, Hybrid/LLM, by question metadata.}
    \label{fig:best-answer-by-metadata}
\end{figure*}
\Cref{fig:best-answer-by-metadata} plots the relation between the winning answer source, LLM or Hybrid, and the question metadata: Working Group, Topic, and difficulty level. Here the LLM answer seems to perform better than the hybrid ones particularly on WGII questions (Mitigation, 28$\%$), on more technical question (Advanced, 29$\%$) and Finance ($32\%$) and Impacts ($27\%$) topics. While hybrid questions are overwhelmingly preferred on the topic of 'Mitigation Pathways' ($97\%$).

\subsection{Rubrics Generation}
\label{sect-app:rubrics}
The following is the model explanation for the generation of the rubric for the question "How does CESM1 Large Ensemble work? What global climate models compose it?"
\paragraph{Model Explanation:}
"The rubric was designed by systematically analyzing the expert preferences. The core of the analysis involved identifying recurring reasons why one answer was preferred over another. This revealed that while clarity and structure were valued, the most critical differentiators for this specific question were factual accuracy regarding the model's composition and methodology.

The key criteria emerged directly from this analysis:

\begin{quote}
    
Explanation of the Large Ensemble Method: This addresses the first part of the question ('How does it work?').

Identification of CESM1 Component Models: This addresses the second part ('What models compose it?').

Accuracy of Technical Details: This criterion was created to capture a crucial point of frequent error identified in the data—specifically, the confusion between the CAM4 and CAM5 atmospheric models. The expert feedback (e.g., preference \#52) highlighted this as a key issue.
Use of Supporting Visuals: Many answers included diagrams, and these were frequently a deciding factor in preferences, often negatively when an image contradicted the text (e.g., showing a CAM4 diagram for a CAM5-based explanation).

The graph summary was highly useful. It helped identify a top tier of answers (10, 4, 3, 9, 7) and a bottom tier (1, 6, 8), providing a general quality ranking. More importantly, the existence of a single large Strongly Connected Component (SCC) indicated cycles and disagreements among raters. This was critical because it showed that simply counting votes was insufficient. For example, Answer 10 had the most votes but contained a significant factual error (citing CAM4). The cycles suggested that different raters weighted criteria differently (e.g., clarity vs. factual accuracy). This insight led to the creation of a rubric that explicitly separates these dimensions, allowing for a more nuanced assessment that could resolve these apparent contradictions. The Pagerank score was also a helpful secondary indicator, highlighting the strength of Answer 7, which my rubric also identified as a top answer.

If I were to do this again, the process was effective. The combination of qualitative reasons, quantitative vote counts, and graph structure provided a rich dataset to create a robust, data-driven rubric."
\end{quote}

%% file: sections/22-appendix-human-experts.tex
\section{Human Experts}
\subsection{Advocates}
\label{sect:app:cf}
We recruited a group of human raters from the Climate Fresk community. Climate Fresk is an NGO, created by Cedric Ringenbach, to facilitate the understanding of climate issues by the general public.\footnote{\url{https://climatefresk.org/world/}.} Specifically, the NGO runs workshops where participants work together to understand the basic concepts of climate change and their causal relations. The material used in the workshops is derived from IPCC reports and participants can continue their journey by becoming facilitators and run workshops themselves. 
The goal behind creating this group was to explore the role of motivation and intrinsic interest in the topic, in combination with a somewhat uniform background on the topic, if not deep academic expertise in the specific topics. It should be noted that the Climate Fresk NGO was not involved in this research project; the raters recruitment was conducted independently.

Participants in the Advocates group were recruited through an interview, based on their Climate Fresk experience as workshop facilitators. 
The group consisted of 17 raters, 8 men and 9 women, from various locations: France (8), Morocco (4), Germany (2), Netherlands (2), India (1). In terms of education: 9 had a PhD, 1 is PhD candidate, 5 had Master's, and 2 Bachelor's degrees. 15 had read one or two IPCC summary for policy makers. 16 1 had read at least parts of the IPCC Tech summary (6 read it completely). All but 1, at least occasionally, read scientific papers and 14 had contributed to some research (e.g., as data analysts), while 10 had authored or co-authored a scientific paper. Climate Fresk experience was assessed by the number of workshops they facilitated: 1-5 workshops (2 raters), 5-10 (1), 10-20 (5), 20+ and certified trainer (6), 20+ and certified trainer instructor (3).

\subsubsection{Quality of Annotations}
\label{sect:app:cf-quality}
\begin{table}[t]
\centering
\caption{Elo scores of candidate answers types, grouped by rater type. Scores are shown for all raters combined, as well as for the 'Experts' and 'Advocates' rater subgroups.}
\label{tab:elo_rater_groups_ranked}
\begin{tabular}{lrrr}
\toprule
System & \multicolumn{1}{c}{All Raters} & \multicolumn{1}{c}{Rated by Experts} & \multicolumn{1}{c}{Rated by Advocates} \\
\midrule
Hybrid curated by Advocates & $\rankone 1044 \pm 23$ & $\rankone 1048 \pm 24$ & $\rankthree 1050 \pm 38$ \\
LLM Answer & $\ranktwo 1033 \pm 16$ & $\ranktwo 1023 \pm 16$ & $\rankone 1066 \pm 29$ \\
Merged Answer& $\rankthree 1011 \pm 16$ & $\rankfour 997 \pm 19$ & $\ranktwo 1062 \pm 26$ \\
Hybrid curated by Experts & $\rankfour 1007 \pm 12$ & $\rankthree 1013 \pm 13$ & $\rankfour 984 \pm 20$ \\
\bottomrule
\end{tabular}
\end{table}

\begin{figure}
    \centering
    \includegraphics[width=0.95\linewidth]{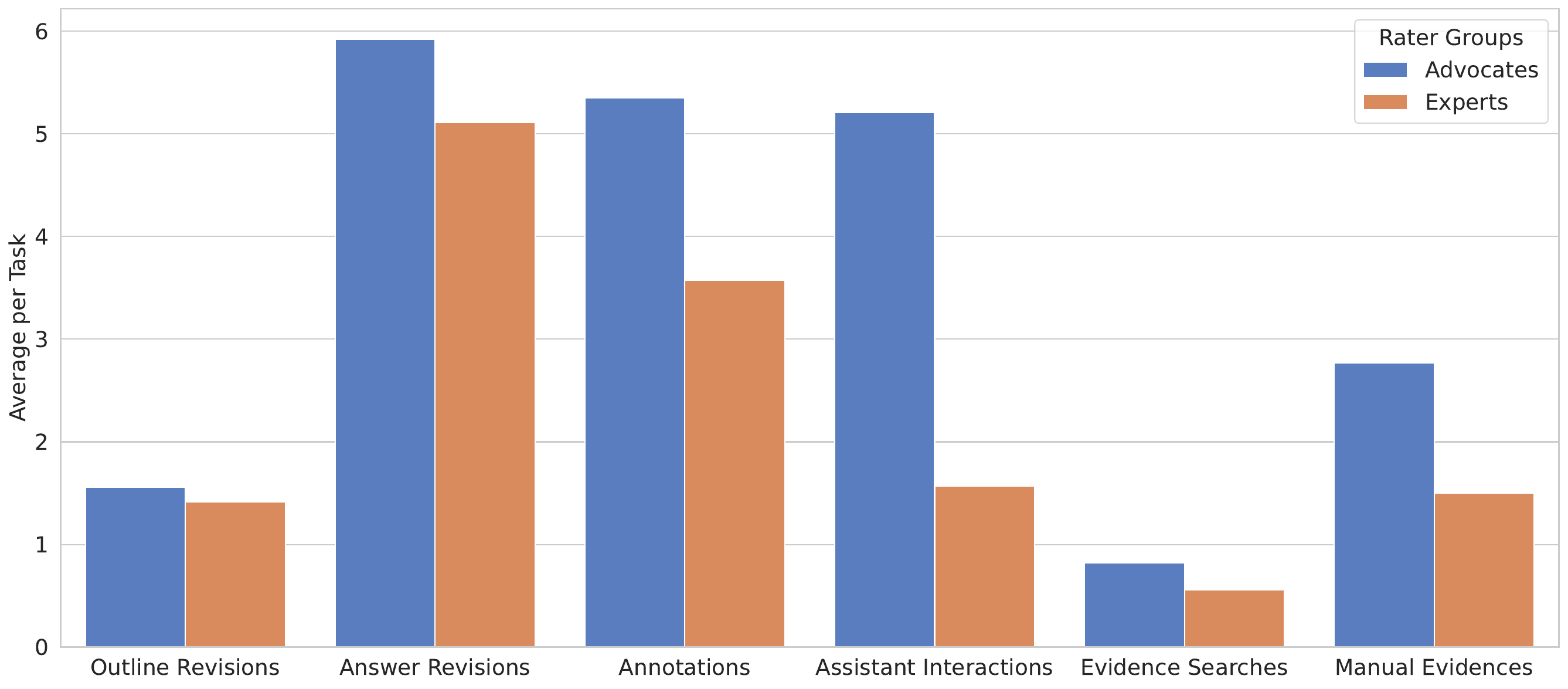}
    \caption{Comparison of Key Interactions in the Editor between Expert groups.}
    \label{fig:editor_interactions}
\end{figure}

\Cref{tab:elo_rater_groups_ranked} shows the scores of model answers, LLM/Merged, or curated either by Advocates (Climate Fresk) or by Experts, and rated by all humans, only Experts or only Advocates. The results show that the Advocates are the best at doing the Phase 1 task as their answers are rated highest by other human participants during Phase 2.  \Cref{fig:editor_interactions} compares the key interactions in the Editor between overall and different expert groups. The data suggests that the Climate Fresk raters are more actively engaged in curating the answer during Phase 1 of the study. They do so through more annotations on the answers, more searches for better evidences, and they manually add more evidences. It seems plausible that through their interactions with the AI-assistant in the Editor, their superior motivation compensates for the different depth of technical knowledge. In the human evaluation of \Cref{tab:results-main}, in only $8\%$ of the battles involving \hybrid{} answers the human-curated answer is from the Advocates group ($92\%$ are from the Experts). However, if the \hybrid{} answer wins $13\%$ of the time the answer has been curated by the Advocates.

At the same time, the Advocates are not as good as the Experts at differentiating the quality between different answers. This can be seen by the close ELO scores between answers as rated by Advocates in \Cref{tab:elo_rater_groups_ranked}. Phase 2 is more difficult to do for participants who don't have the deep knowledge as the Experts do. At the same time, the user interface may be less useful or be more difficult to design for AI assistance, in a side by side.

%% file: sections/24-appendix-ablations.tex
\subsection{\name Autorater Ablations}
\label{sect:app:autorater-ablations}

\begin{table*}[t]
\centering
\footnotesize
\caption{
Elo Score Comparison for the Baseline (CLINB) and ablation Experiments.
}
\label{tab:elo_controls_main}
\begin{tabular}{lrrrrrrr}
\toprule
System & \multicolumn{1}{c}{CLINB} & \multicolumn{1}{c}{-Prompt} & \multicolumn{1}{c}{-Rubrics} & \multicolumn{1}{c}{-Validation} & \multicolumn{1}{c}{-Evidence} & \multicolumn{1}{c}{-Pro+Flash} \\
\midrule
GPT-5 & $\rankone 1157 \pm 19$ & $\ranktwo 1314 \pm 25$ & $\rankone 1185 \pm 20$ & $\ranktwo 1143 \pm 18$ & $\ranktwo 1230 \pm 22$ & $\ranktwo 1096 \pm 17$ \\
Claude Opus 4.1 & $\ranktwo 1144 \pm 17$ & $\rankthree 1094 \pm 20$ & $\ranktwo 1165 \pm 19$ & $\rankone 1145 \pm 18$ & $\rankthree 1087 \pm 21$ & $\rankone 1139 \pm 17$ \\
GPT o3 & $\rankthree 1019 \pm 18$ & $\rankone 1504 \pm 35 $ & $\rankthree 1016 \pm 19$ & $\rankthree 1021 \pm 18$ & $\rankone 1321 \pm 27$ & $\rankthree 1011 \pm 17$ \\
Gemini 2.5 Pro & $\rankfour 968 \pm 18$ & $\rankfour 1037 \pm 21$ & $\rankfour 955 \pm 17$ & $\rankfour 970 \pm 17$ & $\rankfour 1055 \pm 19$ & $\rankfive 962 \pm 16$ \\
Hybrid & $\rankfive 945 \pm 17$ & $\rankseven 567 \pm 31$ & $\rankseven 869 \pm 19$ & $\rankfive 946 \pm 19$ & $\rankseven 749 \pm 24$ & $\rankfour 1010 \pm 18$ \\
Claude Sonnet 4 & $\ranksix 908 \pm 18$ & $\ranksix 740 \pm 21$ & $\rankfive 936 \pm 17$ & $\ranksix 919 \pm 19$ & $\ranksix 777 \pm 22$ & $\ranksix 917 \pm 17$ \\
Gemini 2.5 Flash & $\rankseven 860 \pm 17$ & $\rankfive 744 \pm 23$ & $\ranksix 873 \pm 18$ & $\rankseven 857 \pm 19$ & $\rankfive 781 \pm 22$ & $\rankseven 866 \pm 17$ \\
\bottomrule
\end{tabular}
\end{table*}

\Cref{tab:elo_controls_main} Reports the results of various ablation in the model-assessment procedure.
The baseline (\name) uses the full prompt: guidelines, scientist-reviewed rubrics, evidence handling instructions, step-by-step instructions, and gemini-2.5-pro as the rater.

\paragraph{-Prompt} In this configuration we replace the \name prompt with a simple vanilla  prompt requesting the LLM to choose the better answer. The results change substantially: the human-curated answers get the lower score observed in our experiments. \opus{} experiences also a regression. \geminipro{} and the OpenAI \gpt{} get a boost, with \othree{}'s ELO score increasing by $50\%$. The LLM-as-a-Judge paradigm is known to be susceptible to biases, including favoring style over substance \cite{gudibande2024false} and AI-AI bias \citep{panickssery2024llm}. This probably reveal the underlying bias and stylistic preferences of the judge model and demonstrate the the effectiveness of the \name autorater. 

\paragraph{-Rubrics} Here we remove the question-specific Rubrics components from the \name prompt (see \Cref{sect:eval-prompt} for the full prompt). The ELO scores are close to those obtained with the full \name autorater. However, \hybrid answers drops to last place and the score of the \geminiflash{} improve to overtake the human-curated answers. Notice that \geminiflash and the \hybrid answers are the data from which the rubrics were induced, vias Phase 1 and Phase 2.  This shows that without the question-specific rubrics the model cannot use the detailed information about their relative strengths and weaknesses. The result suggests that data-derived rubrics require more data to be representative of the common concepts, pitfalls and misconception of multiple models and possibly require an adaptive approach where they are constantly updated.

\paragraph{-Validation} uses rubrics generated by \geminipro. These rubrics are generated using experts curated answers and rankings so they already include a lot of expert knowledge. These are the starting rubrics the scientists review and revise into the final rubrics. The scientists were generally very impressed by these rubrics and only made minor improvements during their review so unsurprisingly, the performance of \textit{-Validation} is the same as \name within the confidence interval.

\paragraph{-Evidence} removes the evidence penalty. This is the most crucial component of \name because without it, the autorater cannot tell a valid evidence from an invalid one. It's unable to discount key facts that are unsupported correctly and the scores tend to revert towards the baseline (-Prompt).

\paragraph{-Pro+Flash} swaps out the rating model to \geminiflash. The weaker rating model is not as good at linking multiple pieces of information together from the answer, to the rubrics, and to the associated evidence validity. It is often unable to discount facts backed by invalid evidence so the overall ranking of the models do not agree with \name.

%% file: sections/25-appendix-visuals.tex
\subsection{Visual Evidence}
\label{sect:app:visual}

\begin{figure}
    \centering
    \includegraphics[width=0.95\linewidth]{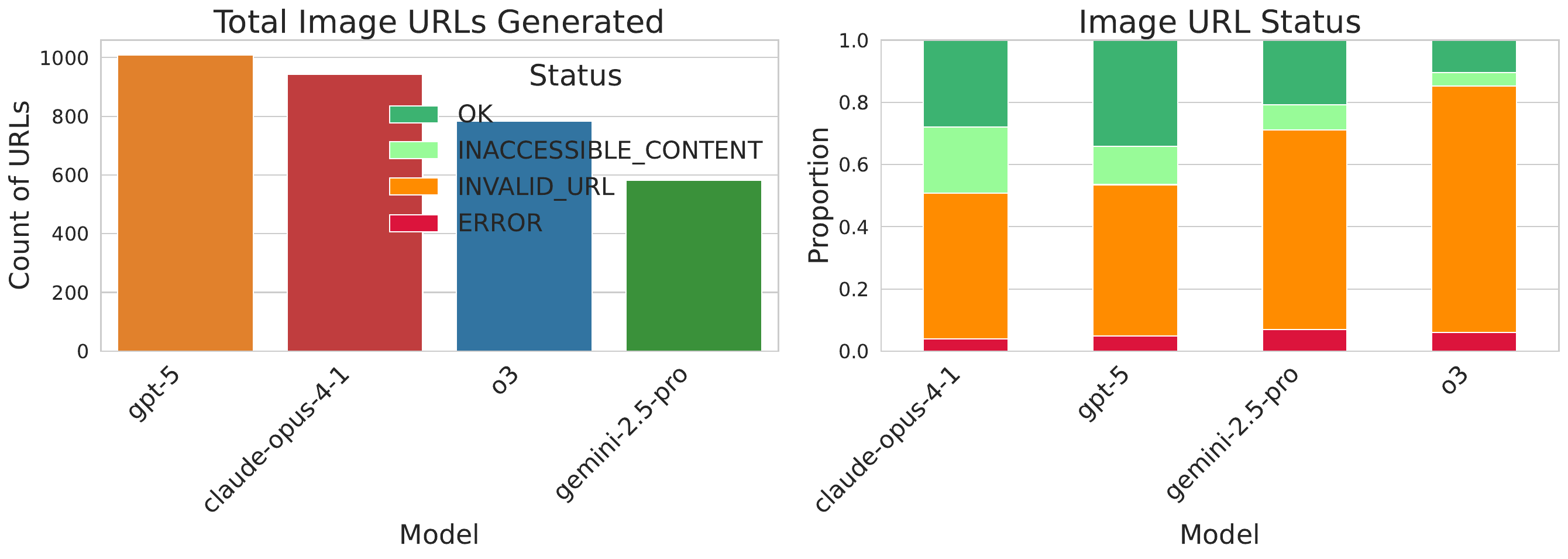}
    \caption{Image URL Status Breakdown for Answers with Mandatory Image}
    \label{fig:image_specific_url_status_control7}
\end{figure}

Most models choose to omit the image in the answer when it is described as optional. When the request to include an image in the response is made "mandatory" most of the image links are hallucinated (\Cref{fig:image_specific_url_status_control7}).

%% file: sections/26-appendix-question-type.tex
\subsection{Question Category and Topics}
\label{sect:app:question-type}
\begin{figure}[htbp]
    \centering 
    
    \begin{subfigure}{0.36\linewidth}
        \centering
        \includegraphics[width=0.95\linewidth]{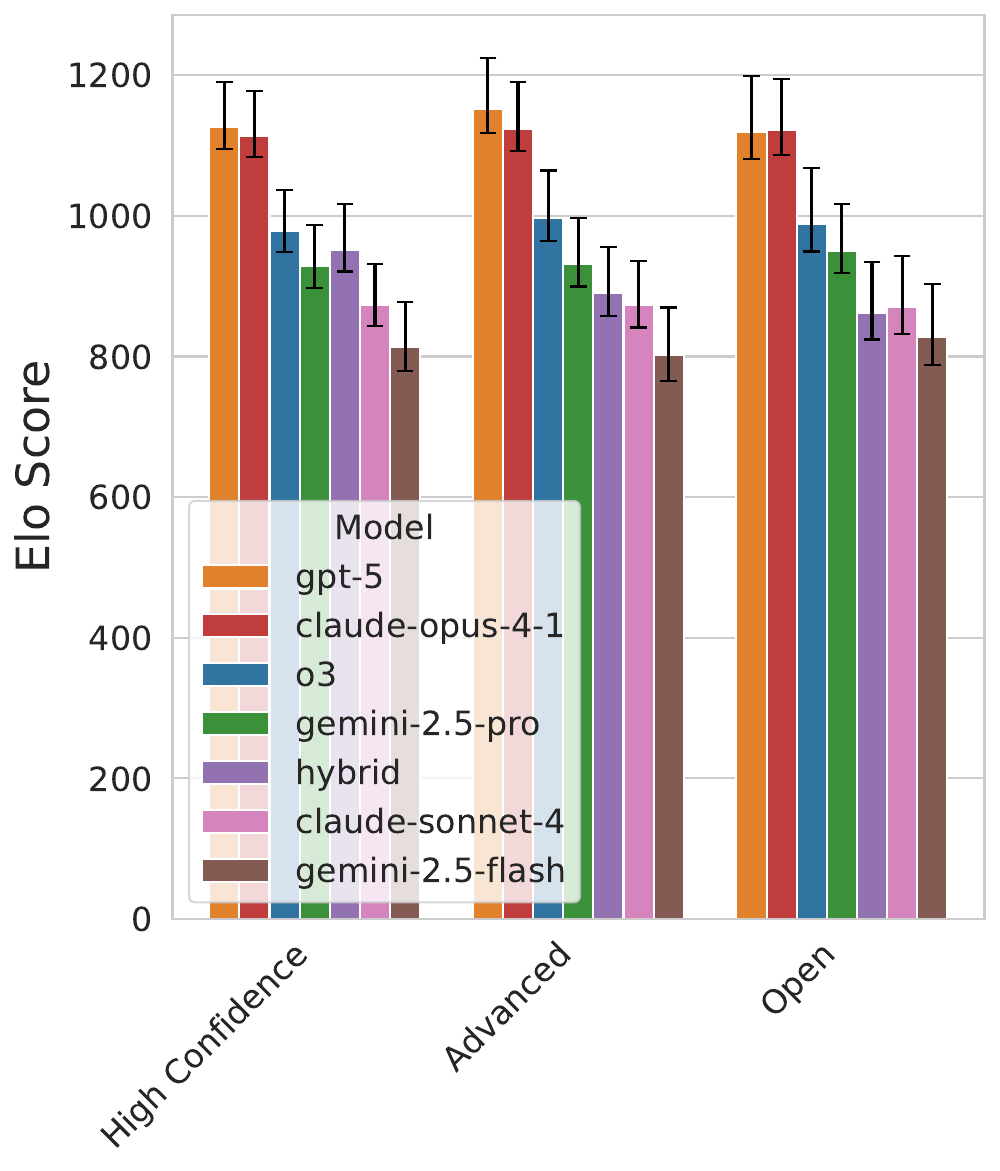}
        \caption{ELO Scores by Question Category}
        \label{fig:elo_scores_by_question_category_clinb}
    \end{subfigure}%
    \hfill 
    \begin{subfigure}{0.62\linewidth}
        \centering
        \includegraphics[width=0.95\linewidth]{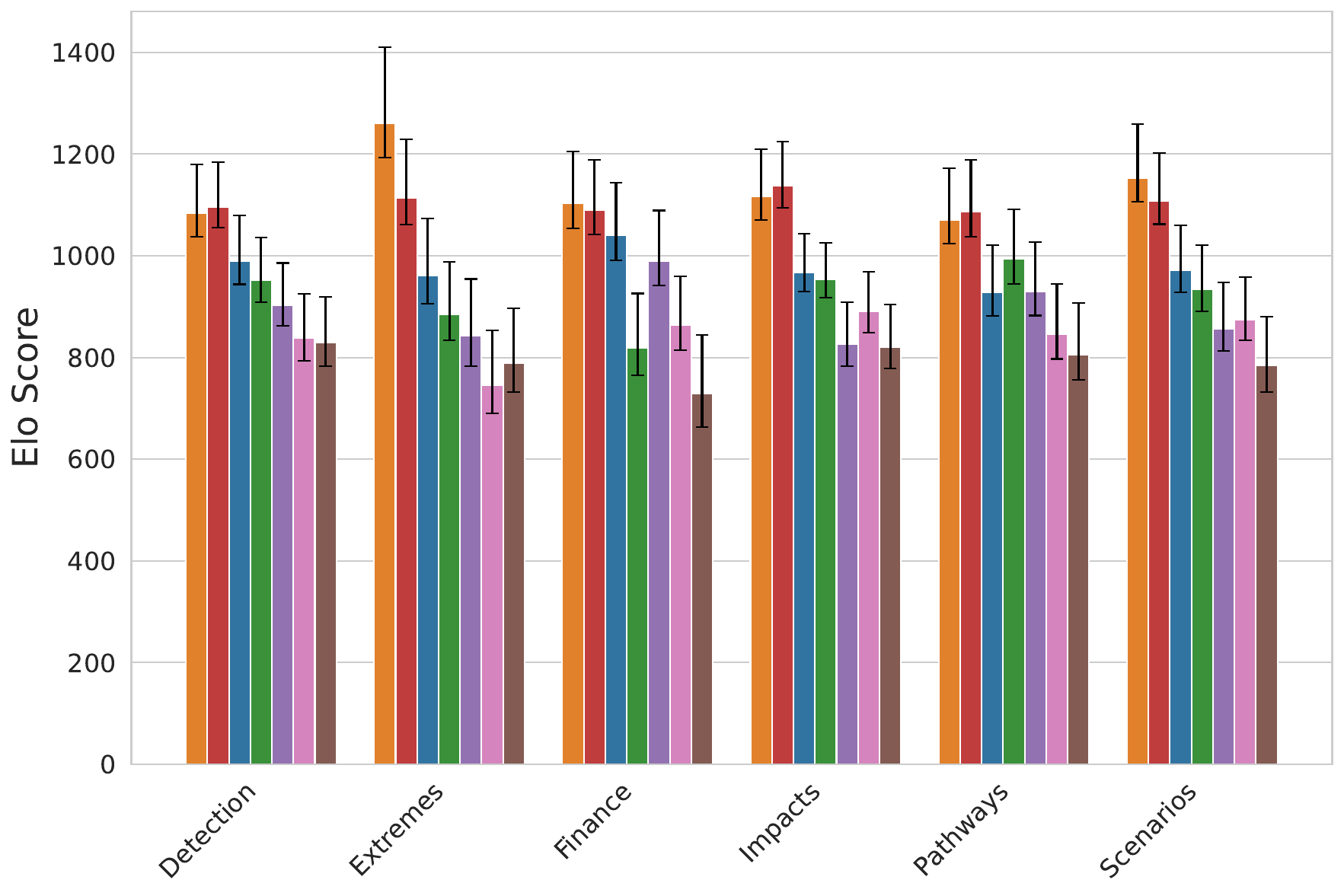}
        \caption{ELO Scores by Question Topic}
        \label{fig:elo_scores_by_question_topic_clinb}
    \end{subfigure}

\end{figure}

In \Cref{fig:elo_scores_by_question_category_clinb}, we look at how well the models answer different questions categories according to the \name autorater: High Confidence, Advanced, and Open. The overall rankings of the models are similar between the 3 groups, and close to the general results, indicating that question complexity does not have a marked effect on answer quality. There are slight variations in performance for the models in the middle. Hybrid answers are slightly better for the High Confidence questions, suggesting humans felt more comfortable improving over the model generated answer for these questions. Perhaps it's because there is a high degree of scientific consensus here. 

In \Cref{fig:elo_scores_by_question_topic_clinb}, we consider how well the models can answer questions on different topics: Detection, Extremes, Finance, Impacts, Pathways, and Scenarios. There is more variation between the different systems when the data is sliced this way. Hybrid answers performed relatively well on "Climate Finance and Risks". Perhaps this means there is higher uncertainty on this topic in the models' parametric knowledge.

%% file: sections/27-appendix-details-eval.tex
\subsection{Detailed Evaluation Dimensions}
\label{sect:app:detailed-dimensions}
\begin{figure}
    \centering
    \includegraphics[width=1\linewidth]{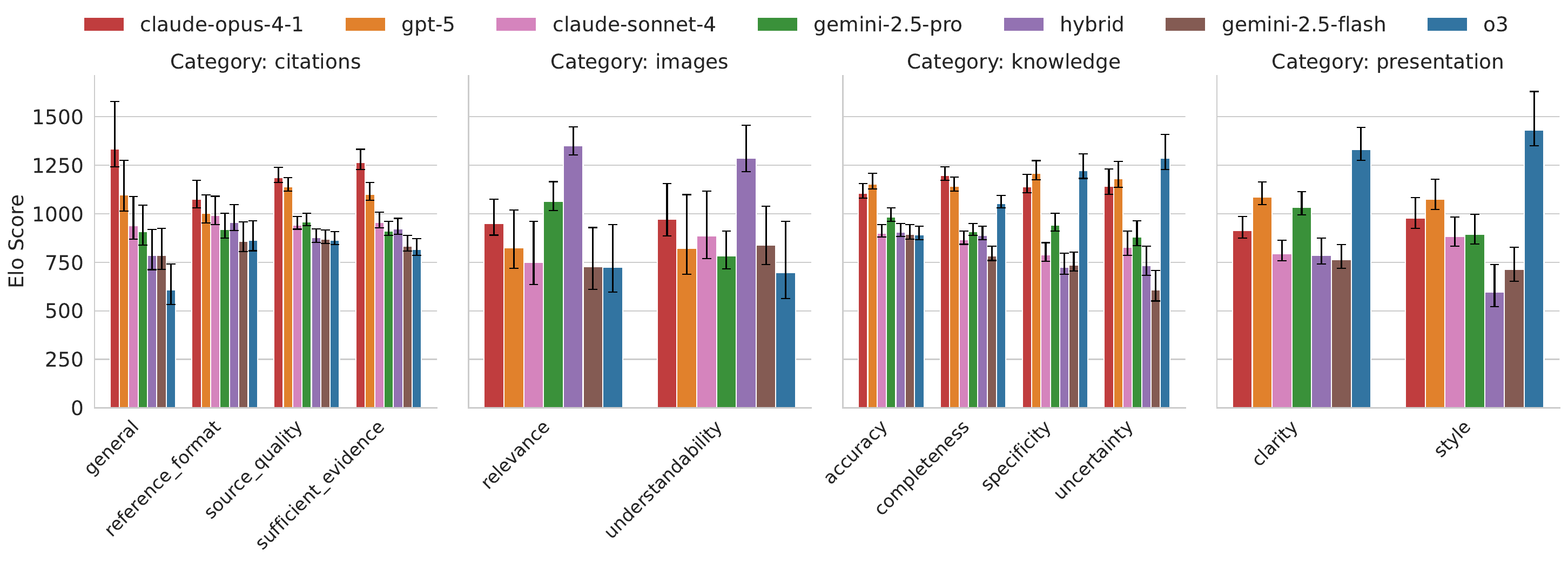}
    \caption{ELO Ratings by Evaluation Dimensions according to CLINB}
    \label{fig:elo_scores_by_dimension_clinb}
\end{figure}
\begin{figure}
    \centering
    \includegraphics[width=1\linewidth]{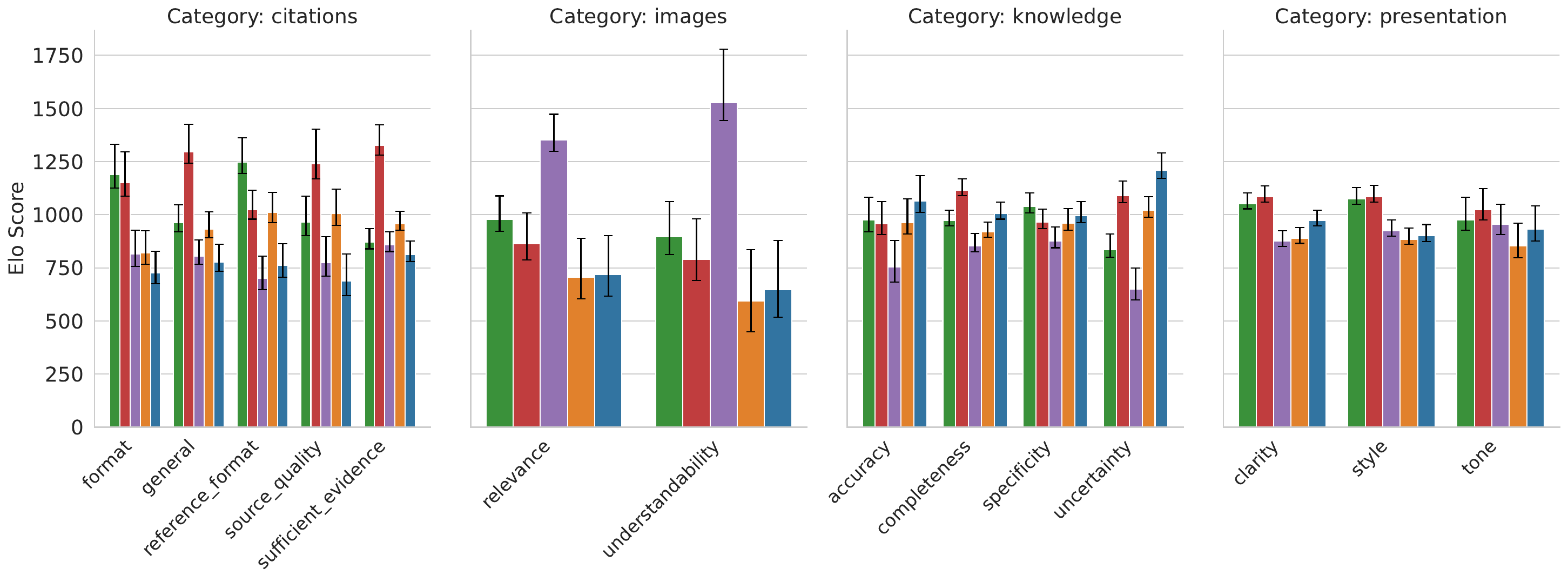}
    \caption{ELO Ratings by Evaluation Dimensions according to Experts}
    \label{fig:elo_scores_by_dimension_experts}
\end{figure}
\Cref{fig:elo_scores_by_dimension_clinb} reports the performance, according to the \name autorater, on a more specific layer of the evaluation dimensions. \opus{} and \gpt{} are head to head on the knowledge dimensions, \opus{} is best on citations, \hybrid{} answers have the best images, while \othree{} wins in presentation.  \Cref{fig:elo_scores_by_dimension_experts} shows the same performance breakdown evaluated by Experts. The experts are not as confident as the \name autorater at judging the knowledge dimensions. We hypothesize that this is due to the fact that exhaustive checking of all technical answer details and nuances is too time consuming for advanced answers. The manual pairwise results show more confidence and resolution on the evidence, images and references, which are easier to check at least superficially.

%% file: sections/28-appendix-prompts.tex
\section{Prompts}
\label{sect:app:prompts}
We report here verbatim the main prompts used in our experiments. Prompts are formatted in Markdown.

\subsection{Guidelines}
\label{sect:guidelines}
These guidelines, derived from~\citep{assessinglargelanguagemodels}, are used both in the autorater prompt and in the system answer prompt, for consistency.

\begin{footnotesize}
\begin{Verbatim}[breaklines=true]
Unless there are significant presentation flaws, disregard stylistic aspects; our goal is to evaluate an answer primarily on its epistemological quality.

In case of ambiguity or underspecification, the question should be interpreted as relevant and specific to the topic of **climate change**.

The answers must adhere strictly to the following guidelines and diagnostic dimensions.

* **Tone:** Answers must maintain a strictly **neutral, balanced, and unbiased** tone. Avoid subjective language, sensationalism, alarmism, exaggeration, or any emotionally loaded phrasing. Focus on objective assessment.
* **Accuracy:** All information provided must be **accurate** according to current scientific understanding and adequate scientific standards. Statements that present scientific findings out of context, are self-contradictory or rely on anecdotal evidence must be avoided.
* **Specificity:** The answer must address **only and exclusively** what the questions explicitly and directly asks. If elaboration is needed, it should deepen the detail on the specific topic rather than broadening the scope to related but irrelevant information. Strictly adhere to any specified context (e.g., temporal or geographical).
* **Completeness:** The answer must address **all parts** of the question thoroughly. Not omit important aspects. Provide sufficient detail, including relevant numbers and statistics where appropriate. Acknowledge established scientific knowledge, relevant findings, and different significant perspectives on the question where applicable.
* **Uncertainty:** The answer must appropriately convey the **degree of uncertainty** associated with the information presented, especially regarding scientific findings or projections. Significant contradicting evidence or alternative interpretations, if they exist within the scientific community, must be mentioned.
* **Sources:**
    * **Every key point, statement, statistic, image or piece of data** presented in the answer must be supported by a traceable citation linked to the reference section.
    * Evidence from **first-hand academic sources**, particularly peer-reviewed articles must be prioritized.
    * The answer must rely only on **trustworthy and authoritative** sources and favor **up-to-date evidence** when newer reliable data or findings are available.
    * All references must always include **valid URLs**.
\end{Verbatim}
\end{footnotesize}

\subsection{System Prompt}
\label{sect:sys-prompt}
This is the prompt used to query the LLMs to be evaluated. Notice that the prompt doesn't specify \emph{how} an answer should be constructed. We argue that it should be up to the system's \emph{intelligence}, i.e., its planning, agentic, self-critiquing, reasoning skills to figure out how to research and compose an adequate response.

\begin{footnotesize}
\begin{Verbatim}[breaklines=true]
**Task:** Carry out the necessary research and answer the question below.

**Contextual Interpretation:** If the question is ambiguous or underspecified, interpret it as being relevant and specific to the topic of **climate change**.

**Input Question:**
{user_question}

**Output Requirements:**

Your answer must strictly adhere to the following format and quality guidelines:

**1. Format:**

* **Main Body:**
    * Maximum length: **{MAX_NUM_WORDS_IN_BENCHMARK_ANSWER} words** (counted by splitting text at whitespaces). Content exceeding this limit will be disregarded.
    * Must be formatted using **Markdown**.

* **Reference Section:**
    * Must list **all** sources cited in the main body.
    * Citations within the text must use the **IEEE style** (e.g., `[1]`, `[2]`).
    * Sources must be numbered sequentially in the order they first appear in the text.
    * Each reference must include a **valid URL link** to the source.
    * Non-academic sources (e.g., web pages, reports) must at least include the title and the source name (e.g., organization, website). Academic sources should follow standard IEEE citation format including authors, year, title, publication venue etc., plus the URL.

* **Images (Optional):**
    * Images can greatly enhance the clarity and effectiveness of the answer. Thus, answers may include images **but only** if they effectively summarize quantitative data, results, describe processes, or significantly aid understanding.
    * Each image must be accompanied by a **caption** describing its content.
    * The caption must **cite the source** of the image, following the same IEEE citation style used in the text, linking to the corresponding entry in the reference section.
    * Each image must include a **valid URL link** directly to the image file itself if possible, or to the page containing the image.
    * Use Markdown for image embedding (`![alt text](Image URL)`). Don't use any other image embedding syntax.

**2. Quality Guidelines:**

The following guidelines must be strictly followed when generating the answer:
{GUIDELINES}

**Proceed to generate the answer based *only* on these instructions.**
\end{Verbatim}
\end{footnotesize}

\subsection{Evaluation Prompt}
\label{sect:eval-prompt}
This is the prompt CLINB uses for pairwise evaluation of answers.
\begin{footnotesize}
\begin{Verbatim}[breaklines=true]
# Task Description

You are an expert climate scientist and internationally known author. Your task is to compare two answers to the same question and provide a detailed assessment of the answers along with a final decision on which answer is better.

You task involves carefully comparing two answers to the same question using the materials listed below:
* The **Question**, marked within <START QUESTION> and <END QUESTION> to be answered.
* **Answer 1**, marked within <START ANSWER 1> and <END ANSWER 1>, the first answer.
* **Answer 2**, marked within <START ANSWER 2> and <END ANSWER 2>, the second answer.
* **Instructions**, marked within <START INSTRUCTIONS> and <END INSTRUCTIONS>, the general instructions on how to perform the task, decide which answer is better, and why, in a principled and structured way.
* **Guidelines**, marked within <START GUIDELINES> and <END GUIDELINES>, the primary evaluation dimensions.
* **Grader's Cheat Sheet**, marked within <START CHEAT SHEET> and <END CHEAT SHEET>, the question-specific background information that is necessary for an accurate and consistent evaluation.
* **Question-specific Grading Rubrics**, marked within <START RUBRICS> and <END RUBRICS>, the answer grading standards that are specific to the question.
* **Shared Grading Rubrics**, marked within <START SHARED RUBRICS> and <END SHARED RUBRICS>, the grading standards for the answers, that are shared across multiple questions.
* **Supplemental Materials**, marked within <START SUPPLEMENTAL MATERIALS> and <END SUPPLEMENTAL MATERIALS>, the supplemental materials that are used for the evaluation.

# Main Task Materials
<START QUESTION>
[The question is inserted here]
<END QUESTION>

<START ANSWER 1>
[answer 1 is inserted here]
<END ANSWER 1>

<START ANSWER 2>
[answer 2 is inserted here]
<END ANSWER 2>

<START INSTRUCTIONS>
# Step-by-step Instructions

## **Step 1: Analyze the Question and the Answer Contents**

First, get familiar with the question and content of both answers.

## **Step 2: Analyze the Grading Rubric**

The question-specific Grading Rubric is marked within <START RUBRICS> and <END RUBRICS>.
The shared Grading Rubric is marked within <START SHARED RUBRICS> and <END SHARED RUBRICS>.

The Grading Rubric has been created along the following core assessment criteria:
* **Scientific Accuracy and Depth:** Correctness and thoroughness of the climate science concepts.
* **Clarity of Argument and Structure:** Logical flow, coherence, and clear writing.
* **Use of Evidence (Images & Data):** Effective and accurate integration of images to support the argument.
* **Quality of Citations and References:** Appropriate use of high-quality sources.
* **Adherence to the 'Guidelines':** How well the answer follows the specified quality guidelines.

In addition to the question-specific Grading Rubric, you will also be provided with the shared Grading Rubric, marked within <START SHARED RUBRICS> and <END SHARED RUBRICS>.
This shared Grading Rubric is an additional set of quality criteria for the answers. However pay most attention to the question-specific Grading Rubric and the Grader's Cheat Sheet.

You should analyze both the question-specific Grading Rubric and the shared Grading Rubric, with an **emphasis on the question-specific Grading Rubric**.

## **Step 3: Analyze the Grader's Cheat Sheet**

The Grader's Cheat Sheet is marked within <START CHEAT SHEET> and <END CHEAT SHEET>.
It can help you apply the rubric quickly and consistently.
It includes the following sections:
* **Key Scientific Concepts to Look For:** A bulleted list of the essential scientific points a top answer must include.
* **Common Pitfalls & Misconceptions:** A list of frequent errors or weaker arguments.
* **Checklist for Evaluation of Images and References:** 2-3 questions that guide the assessment of images and references.

## **Step 4: Analyze the Answer Guidelines**

Analyze the provided Answer Guidelines, which is a set of quality criteria for answers.

## **Step 5: Analyze and Grade Each Answer Individually.**

Meticulously analyze each answer's content with respect to the above Grading Rubrics and the Grader's Cheat Sheet.
Make this analysis in depth and as precise as possible and reflect on your choices.
Then go through the rubrics one by one and, for each one, provide an assessment and a rating for both answers.
In addition, for each criterion, compare the two answers and decide if one answer is better than the other.
It is possible for both answer to have the same score but one can still have an edge over the other one.
Use 'answer_1_preferred' to indicate that answer_1 is better than answer_2 and 'answer_2_preferred' to indicate that answer_2 is better than answer_1.
If there is no clear winner, declare a 'tie'.

Consider the following guidelines when rating the answers:
* **Focus on the Question-Specific Grading Rubric:** Prioritize the question-specific Grading Rubric and the Grader's Cheat Sheet over the shared Grading Rubric and the Answer Guidelines.
* **Provide Detailed and Scientific Rationales:** Provide a detailed and scientific rationale for your ratings, including supporting evidence from the answer and the Grading Rubric.
* **Be Critical:** Be thorough and rigorous in your analysis and ratings.

## **Step 6: Make a Final Decision**

Finally, make a final decision on which answer is better based on your previous analysis and the available materials.
Give a short explanation of what aspects of the answers are the main drivers for your decision.

Use the Supplemental Materials to determine if any penalty should be applied to the answer.

Explain why the preferred answer is better than the other one based on the grading materials.
If an answer contains images pay particular attention to them and how they contibute to the answer.

## Guidelines
In your assessment, follow strictly the evaluation dimensions explained in the guidelines below:
<START GUIDELINES>
[The guidelines (above) are inserted here]
<END GUIDELINES>

## Grader's Cheat Sheet
Use the Cheat Sheet to help you understand the question and the background information necessary to formulate a calibrated and consistent assessment.
<START CHEAT SHEET>
[Question specific cheat sheet is inserted here]
<END CHEAT SHEET>

## Specific Grading Rubrics
Use the Grading Rubrics to implement consistent and accurate grading standards.
<START RUBRICS>
[Question specific rubrics are inserted here]
<END RUBRICS>

## Shared Grading Rubrics
Use the Shared Grading Rubrics to implement consistent and accurate grading standards.
<START SHARED RUBRICS>
[
  {
    "criterion": "Tone",
    "description": "Evaluates the neutrality, balance, and objectivity of the language used in the answer.",
    "requirements_score_9_10": "The answer maintains a consistently neutral, balanced, and unbiased tone throughout. It is entirely free of subjective language, sensationalism, alarmism, exaggeration, or emotionally loaded phrasing. The focus is purely on objective, factual assessment.",
    "requirements_score_7_8": "The tone is overwhelmingly neutral and objective, but there may be very minor, isolated instances of slightly subjective or subtly leading language that do not significantly impact the overall balance of the answer.",
    "requirements_score_5_6": "The answer attempts to maintain a neutral tone, but there are noticeable lapses into subjective, slightly exaggerated, or emotionally tinged language in several places, which moderately detracts from the overall objectivity.",
    "requirements_score_1_4": "The tone is frequently and significantly biased, subjective, sensationalist, or alarmist. Emotionally loaded language is common and severely compromises the objectivity and neutrality of the answer.",
    "requirements_score_0": "The tone is entirely inappropriate, propagandistic, or completely fails to adopt a factual and objective stance. The answer is unrateable on its epistemological merits due to its tone."
  },
  {
    "criterion": "Accuracy",
    "description": "Assesses the factual correctness of the information provided, its contextual relevance, and its basis in current scientific understanding.",
    "requirements_score_9_10": "All information presented is entirely accurate according to current, mainstream scientific understanding. All scientific findings are presented in their proper context, without contradictions or reliance on anecdotal evidence. The information reflects the highest scientific standards.",
    "requirements_score_7_8": "The vast majority of the information is accurate and well-contextualized. There may be a minor factual error or a slightly out-of-context statement that does not undermine the core argument or the overall correctness of the answer.",
    "requirements_score_5_6": "The answer contains some accurate information but also includes several noticeable factual errors, misinterpretations, or statements presented out of their proper scientific context. The answer may contain minor self-contradictions.",
    "requirements_score_1_4": "The answer contains significant, numerous, or fundamental factual errors. Information is frequently presented out of context, is self-contradictory, or relies on discredited science or anecdotal evidence. The core claims are scientifically unsound.",
    "requirements_score_0": "The answer is completely factually incorrect, based on pseudoscience or misinformation, or provides no verifiable information."
  },
  {
    "criterion": "Specificity",
    "description": "Evaluates how strictly the answer adheres to the scope of the question, including any explicit constraints (e.g., temporal or geographical).",
    "requirements_score_9_10": "The answer addresses only and exclusively the explicit question asked. It does not introduce related but irrelevant topics. Any elaboration serves only to deepen the detail on the specific topic. All specified contexts (e.g., timeframe, location) are strictly respected.",
    "requirements_score_7_8": "The answer is highly focused on the question but may include a minor, brief deviation into a tangentially related topic that does not significantly detract from the main focus. It largely respects all specified contexts.",
    "requirements_score_5_6": "The answer addresses the core question but also includes a noticeable amount of irrelevant information or broadens the scope beyond what was asked. It may partially neglect or misinterpret a specified context.",
    "requirements_score_1_4": "The answer significantly deviates from the question, focusing largely on related but irrelevant topics. The core question is only superficially addressed or misunderstood. Specified contexts are largely ignored.",
    "requirements_score_0": "The answer completely fails to address the question asked and is entirely off-topic."
  },
  {
    "criterion": "Completeness",
    "description": "Assesses whether the answer thoroughly addresses all parts of the question, providing sufficient detail and acknowledging all relevant perspectives.",
    "requirements_score_9_10": "The answer comprehensively addresses all parts of the question, leaving no significant aspect unexamined. It provides a thorough level of detail, including relevant and well-chosen data and statistics. It accurately represents the established scientific consensus and acknowledges different significant scientific perspectives where applicable.",
    "requirements_score_7_8": "The answer addresses all major parts of the question but may omit a minor aspect or provide slightly less detail than would be ideal in one area. The overall picture is still robust and well-supported.",
    "requirements_score_5_6": "The answer addresses the main parts of the question but omits some important aspects, fails to provide sufficient detail, or lacks nuance.",
    "requirements_score_1_4": "The answer is notably incomplete, omitting major parts of the question or providing only superficial, cursory information. Key details, data, or established scientific perspectives are missing.",
    "requirements_score_0": "The answer is a fragment or completely fails to provide a substantive response to any part of the question."
  },
  {
    "criterion": "Sources",
    "description": "Assesses the quality, appropriateness, and proper citation of the sources used to support the answer.",
    "requirements_score_9_10": "Every key point, statement, statistic, image or piece of data is meticulously supported by a traceable citation linked to the reference section. Sources are overwhelmingly first-hand, peer-reviewed academic literature or reports from major scientific bodies (e.g., IPCC, Nature, WHO). All sources are authoritative, up-to-date, and include a valid, working URL.",
    "requirements_score_7_8": "Most key points are properly cited with high-quality, authoritative sources. There may be a few minor statements lacking a direct citation, or a small number of citations may be to high-quality secondary sources (e.g., reputable scientific journalism or a good Wikipedia article) instead of primary literature. All URLs are valid.",
    "requirements_score_5_6": "Citations are present but inconsistent; some key claims lack support. The answer relies significantly on secondary sources, news articles, or institutional websites rather than primary academic literature. Some sources may be slightly outdated.",
    "requirements_score_1_4": "Citations are largely missing, incorrect, or link to unreliable sources (e.g., blogs, opinion pieces, non-scientific websites). There is little to no use of peer-reviewed literature. Many key claims are unsupported.",
    "requirements_score_0": "The answer provides no sources or citations whatsoever, or the sources provided are entirely inappropriate and untrustworthy."
  }
]
<END SHARED RUBRICS>

## Supplemental Materials
Use the Supplemental Materials to determine if the referenced images and URLs are valid and accessible. If an URL is not explicitly included in the Supplemental Materials, assume it is valid. The answer should be penalized if it contains invalid images or URLs.
If an image is indicated as important for answering the question but it is missing in the answer being evaluated, the answer should also be penalized.
Assume that references are important for answering the question. If there are no references in the answer being evaluated, the answer should be penalized. Also, the answer with more valid references should be rewarded.

If an answer contains invalid references, you must assume that the facts supported by those references are missing from the answer as they cannot be verified.
If key facts in the answer are not supported by a valid reference, you must also assume those facts are missing from the answer as they cannot be verified.
This will affect how you evaluate the Completeness, Accuracy, and Specificity of the answer.
<START SUPPLEMENTAL MATERIALS>
[Answer specific supplemental materials are inserted here. These includes the results of checking the validity of each referenced URL.]
<END SUPPLEMENTAL MATERIALS>
<END INSTRUCTIONS>

Output format is JSON:
{
  "specific_rubric_ratings: [  # The specific rubric ratings as given above between <START RUBRICS> and <END RUBRICS>.
      {
        criterion: str # The name of the criterion.
        eval_answer_1: str  # Short explanation of how answer 1 scores against the rubric.
        rating_answer_1: float  # Float (with single decimal space) from 0 to 10
        eval_answer_2: str  # Short explanation of how answer 2 scores against the rubric.
        rating_answer_2: float  # Float (with single decimal space) from 0 to 10
        pair_eval_rationale: str # Rationale for the pair evaluation of answer 1 vs answer 2.
        pair_eval: str  # Either "answer_1_preferred", "answer_2_preferred", or "tie"
      },
  ],
  "shared_rubric_ratings": [  # The shared rubric ratings as given above between <START SHARED RUBRICS> and <END SHARED RUBRICS>.
      {
        criterion: str # The name of the criterion.
        eval_answer_1: str  # Short explanation of how answer 1 scores against the rubric.
        rating_answer_1: float  # Float (with single decimal space) from 0 to 10
        eval_answer_2: str  # Short explanation of how answer 2 scores against the rubric.
        rating_answer_2: float  # Float (with single decimal space) from 0 to 10
        pair_eval_rationale: str # Rationale for the pair evaluation of answer 1 vs answer 2.
        pair_eval: str  # Either "answer_1_preferred", "answer_2_preferred", or "tie"
      },
  ],
  "decision": str, # Either "answer_1_preferred", "answer_2_preferred", no ties allowed here.
  "explanation": str # The rationale behind the decision.
}
Output:
\end{Verbatim}
\end{footnotesize}

%% file: sections/29-appendix-characters.tex
\section{Informal Characterizations}
\label{app:characters}
\textbf{\gpt's} answers are rated as accurate, complete, specific, nuanced, up to date, and of great scientific depth.
Regarding presentation they are logically organized and often include some advanced concepts, that others models omit. They tend to be text-heavy though and light on helpful formatting.

\textbf{\othree} is praised for scientific nuance and completeness. \othree is more likely to give details, cite numbers and direct facts. On the flipside this opens up the answer to being factually wrong for each of these facts. The model also seems to have more problems with hallucinated references than other models. Another strength is presentation, in particular using tables, formatting, and a clear logical flow. As an example for accessible formatting, for the question "Can you detail the assumptions in the five different climate scenario SSP1-1.9, SSP1-2.6, SSP2-4.5, SSP3-7.0 and SSP5-8.5.", \othree produced a clear table allowing for easy side-by-side comparison of the different scenarios. This was preferred over the narrative styles of the other models, particularly \opus.

Our analysis shows that \textbf{\opus} answers not only rate among the best for scientific depth, comprehensiveness, specificity, and nuance but also provide verifiable data and quantitative detail, backed by extensive references. The main weakness seems to lie in interpretation of the question: sometimes \opus misinterprets the question and answers something else.
For example, for a question about impact attribution studies of extreme weather events, \opus fails to define \emph{impact attribution}, i.e. how climate change influences the \emph{outcomes} of an extreme weather event, and instead produces a comprehensive answer on event attribution, i.e. quantifying how climate change affects the likelihood of the extreme weather event \emph{itself}.

According to the analysis, \textbf{\geminipro's} strength lies in providing reliable and accurate answers backed by high-quality sources. The model seems to be good at correctly interpreting the question, identifying the core scientific principles and giving a direct and correct answer. Compared to other models, \geminipro answers tend to be less extensive and can lack some nuance, completeness, or concrete numbers.

Finally, the main strengths for the \textbf{\hybrid} answers are verifiability and presentation. These answers often win because all references can be resolved, in contrast to other systems, as humans are immune to hallucinating URLs. In addition, most \hybrid answers include helpful images and graphs (see Figure \ref{fig:evidence-status}, see \cref{app:examples}). On the flip side, \hybrid answers can be superficial and lack some concrete numbers or examples (also \cref{app:examples}).

%% file: sections/30-appendix-examples.tex
\section{Examples}
\label{app:examples}

In this section we investigate an example of a disagreement between the human preferences and our rubrics based autorater. The goal is to gain a high-level understanding of potential biases humans or prompted LLMs might exhibit, not to second guess the human rating to align with the automatic ones. We believe that there is value in having a multifaceted assessment and thus focus on the correctness of the justification over the individual preferences.

\subsection{Example 1: What kind of data and measurement tool is best for climate risk modeling and disclosure}

\paragraph{Question} \textit{What kind of data and measurement tool is best for climate risk modeling and disclosure?}

\paragraph{Summary of Verdicts} Scientists preferred the model answer, citing superior presentation clarity and style. The Autorater preferred the Hybrid (Human + Gemini Flash) answer.

\paragraph{Analysis} The Autorater's preference for the Hybrid answer highlights the importance of substantive completeness as defined by the expert-curated rubric.
\begin{enumerate}
\item \textbf{Addressing Uncertainty and Challenges:} The rubric explicitly required a sophisticated discussion of the inherent challenges and uncertainties in climate risk modeling. The Autorater found the Hybrid answer vastly superior on this dimension. The human expert intervention ensured this critical aspect was covered comprehensively, while the model response addressed them only superficially.
    \item \textbf{Citation and Image Quality:} The Autorater penalized the model answer for including invalid references supporting critical frameworks (NGFS and IFRS S2) and for providing a broken image link.
\end{enumerate}
This divergence demonstrates how the Autorater, guided by the rubric, prioritizes essential scientific content—such as the acknowledgment of uncertainty—over stylistic clarity, mitigating potential presentation bias in evaluation. Furthermore, this case illustrates the value of the expert-in-the-loop approach, where human curation ensured substantive completeness even when starting from a less capable base model (\geminiflash) compared to a stronger autonomous model.

\subsubsection{Rubrics}
See Tables \cref{tab:ex2-rubrics,tab:ex2-shared-rubrics-1,tab:ex2-shared-rubrics-2} for the specific and shared rubrics used.

\begin{table}[htp]
\footnotesize
\centering
\caption{Question specific rubrics, reviewed and edited by \emph{scientists}.}
\label{tab:ex2-rubrics}
\begin{tabular}{lp{11cm}}
\toprule

\textbf{Question} & What kind of data and measurement tool is best for climate risk modeling and disclosure? \\
\midrule
\noalign{\vskip 1mm}    
\multicolumn{2}{l}{\textbf{Risk Categorization and Scope}} \\
\noalign{\vskip 2mm}    
description &  Assesses the clarity and accuracy of the distinction between physical and transition risks, and the breadth of the discussion to cover various sectors, not just finance. \\
score 9-10 &  Clearly and accurately defines and distinguishes between physical risks (both acute and chronic) and transition risks (policy, technology, market). The discussion is comprehensive and not limited to a single sector like finance. \\
score 7-8 &  Correctly defines and distinguishes physical and transition risks, but may lack detail on their sub-types. The scope is generally appropriate but might lean heavily towards one sector. \\
score 5-6 &  Mentions both risk types, but the distinction is unclear or the definitions are superficial. The scope is likely too narrow (e.g., only finance). \\
score 1-4 &  Fails to clearly distinguish between risk types, discusses only one type, or the definitions are incorrect. \\
score 0 &  Does not address the different categories of climate risk. \\
\midrule
\noalign{\vskip 1mm}    
\multicolumn{2}{l}{\textbf{Specificity of Data and Tools}} \\
\noalign{\vskip 2mm}    
description &  Evaluates the level of detail in describing the specific data types, models, and analytical techniques for both physical and transition risk modeling. \\
score 9-10 &  Provides a rich, specific list of data types and tools for both risk categories. Mentions specific model types (e.g., GCMs/RCMs, catastrophe models), techniques (e.g., downscaling, scenario analysis), and data sources (e.g., geospatial data, policy trackers). \\
score 7-8 &  Lists relevant data types and tools for both risk categories but with less specific detail. For example, might mention 'climate models' without specifying GCMs, or 'financial models' without mentioning scenario analysis. \\
score 5-6 &  Describes data and tools in general terms (e.g., 'climate data,' 'modeling'). The link between specific risks and the corresponding tools is weak or absent. \\
score 1-4 &  Mentions data or tools, but the description is vague, inaccurate, or highly incomplete. \\
score 0 &  Fails to mention relevant data or measurement tools. \\
\midrule
\noalign{\vskip 1mm}    
\multicolumn{2}{l}{\textbf{Discussion of Frameworks and Standards}} \\
\noalign{\vskip 2mm}    
description &  Assesses the inclusion and explanation of key industry and regulatory frameworks that guide climate risk disclosure and assessment. \\
score 9-10 &  Explicitly names and accurately describes the role of key disclosure frameworks (e.g., TCFD, ISSB) and may also mention data standards (e.g., GHG Protocol). Effectively uses or describes a conceptual assessment framework to structure the process. \\
score 7-8 &  Names relevant frameworks (e.g., TCFD) but provides a limited explanation of their role or significance. May include a conceptual framework diagram but with minimal integration into the text. \\
score 5-6 &  Mentions frameworks in passing or lists them without any explanation of their purpose. \\
score 1-4 &  Fails to mention key frameworks or shows a misunderstanding of their purpose. \\
score 0 &  No mention of any frameworks or standards. \\
\midrule
\noalign{\vskip 1mm}    
\multicolumn{2}{l}{\textbf{Nuance and Acknowledgment of Challenges}} \\
\noalign{\vskip 2mm}    
description &  Evaluates the answer's acknowledgment of the inherent complexities, uncertainties, and limitations in climate risk modeling and disclosure. \\
score 9-10 &  Provides a sophisticated discussion of key challenges, such as data availability and granularity, model limitations, inherent uncertainty in climate projections, and issues with the transparency and comparability of tools. \\
score 7-8 &  Acknowledges that uncertainties and challenges exist, but the discussion is less detailed. May mention 'data gaps' or 'model uncertainty' without significant elaboration. \\
score 5-6 &  Makes a brief, generic statement about uncertainty without connecting it to specific aspects of climate risk modeling. \\
score 1-4 &  Ignores the topic of uncertainty and challenges, presenting information as if it were perfectly known and straightforward. \\
score 0 &  The answer is too brief or irrelevant to assess this criterion. \\
\bottomrule
\end{tabular}
\end{table}

\begin{table}[htp]
\footnotesize
\centering
\caption{Shared (non question specific) rubrics used for evaluation (Part 1).}
\label{tab:ex2-shared-rubrics-1}
\begin{tabular}{lp{11cm}}
\toprule
\noalign{\vskip 1mm}    
\multicolumn{2}{l}{\textbf{Tone}} \\
\noalign{\vskip 2mm}    
description &  Evaluates the neutrality, balance, and objectivity of the language used in the answer. \\
score 9-10 &  The answer maintains a consistently neutral, balanced, and unbiased tone throughout. It is entirely free of subjective language, sensationalism, alarmism, exaggeration, or emotionally loaded phrasing. The focus is purely on objective, factual assessment. \\
score 7-8 &  The tone is overwhelmingly neutral and objective, but there may be very minor, isolated instances of slightly subjective or subtly leading language that do not significantly impact the overall balance of the answer. \\
score 5-6 &  The answer attempts to maintain a neutral tone, but there are noticeable lapses into subjective, slightly exaggerated, or emotionally tinged language in several places, which moderately detracts from the overall objectivity. \\
score 1-4 &  The tone is frequently and significantly biased, subjective, sensationalist, or alarmist. Emotionally loaded language is common and severely compromises the objectivity and neutrality of the answer. \\
score 0 &  The tone is entirely inappropriate, propagandistic, or completely fails to adopt a factual and objective stance. The answer is unrateable on its epistemological merits due to its tone. \\
\midrule
\noalign{\vskip 1mm}    
\multicolumn{2}{l}{\textbf{Accuracy}} \\
\noalign{\vskip 2mm}    
description &  Assesses the factual correctness of the information provided, its contextual relevance, and its basis in current scientific understanding. \\
score 9-10 &  All information presented is entirely accurate according to current, mainstream scientific understanding. All scientific findings are presented in their proper context, without contradictions or reliance on anecdotal evidence. The information reflects the highest scientific standards. \\
score 7-8 &  The vast majority of the information is accurate and well-contextualized. There may be a minor factual error or a slightly out-of-context statement that does not undermine the core argument or the overall correctness of the answer. \\
score 5-6 &  The answer contains some accurate information but also includes several noticeable factual errors, misinterpretations, or statements presented out of their proper scientific context. The answer may contain minor self-contradictions. \\
score 1-4 &  The answer contains significant, numerous, or fundamental factual errors. Information is frequently presented out of context, is self-contradictory, or relies on discredited science or anecdotal evidence. The core claims are scientifically unsound. \\
score 0 &  The answer is completely factually incorrect, based on pseudoscience or misinformation, or provides no verifiable information. \\
\midrule
\noalign{\vskip 1mm}    
\multicolumn{2}{l}{\textbf{Specificity}} \\
\noalign{\vskip 2mm}    
description &  Evaluates how strictly the answer adheres to the scope of the question, including any explicit constraints (e.g., temporal or geographical). \\
score 9-10 &  The answer addresses only and exclusively the explicit question asked. It does not introduce related but irrelevant topics. Any elaboration serves only to deepen the detail on the specific topic. All specified contexts (e.g., timeframe, location) are strictly respected. \\
score 7-8 &  The answer is highly focused on the question but may include a minor, brief deviation into a tangentially related topic that does not significantly detract from the main focus. It largely respects all specified contexts. \\
score 5-6 &  The answer addresses the core question but also includes a noticeable amount of irrelevant information or broadens the scope beyond what was asked. It may partially neglect or misinterpret a specified context. \\
score 1-4 &  The answer significantly deviates from the question, focusing largely on related but irrelevant topics. The core question is only superficially addressed or misunderstood. Specified contexts are largely ignored. \\
score 0 &  The answer completely fails to address the question asked and is entirely off-topic. \\

\bottomrule
\end{tabular}
\end{table}

\begin{table}[htp]
\footnotesize
\centering
\caption{Shared (non question specific) rubrics used for evaluation (Part 2).}
\label{tab:ex2-shared-rubrics-2}
\begin{tabular}{lp{11cm}}
\toprule
\noalign{\vskip 1mm}    
\multicolumn{2}{l}{\textbf{Completeness}} \\
\noalign{\vskip 2mm}    
description &  Assesses whether the answer thoroughly addresses all parts of the question, providing sufficient detail and acknowledging all relevant perspectives. \\
score 9-10 &  The answer comprehensively addresses all parts of the question, leaving no significant aspect unexamined. It provides a thorough level of detail, including relevant and well-chosen data and statistics. It accurately represents the established scientific consensus and acknowledges different significant scientific perspectives where applicable. \\
score 7-8 &  The answer addresses all major parts of the question but may omit a minor aspect or provide slightly less detail than would be ideal in one area. The overall picture is still robust and well-supported. \\
score 5-6 &  The answer addresses the main parts of the question but omits some important aspects, fails to provide sufficient detail, or lacks nuance. \\
score 1-4 &  The answer is notably incomplete, omitting major parts of the question or providing only superficial, cursory information. Key details, data, or established scientific perspectives are missing. \\
score 0 &  The answer is a fragment or completely fails to provide a substantive response to any part of the question. \\
\midrule
\noalign{\vskip 1mm}    
\multicolumn{2}{l}{\textbf{Sources}} \\
\noalign{\vskip 2mm}    
description &  Assesses the quality, appropriateness, and proper citation of the sources used to support the answer. \\
score 9-10 &  Every key point, statement, statistic, image or piece of data is meticulously supported by a traceable citation linked to the reference section. Sources are overwhelmingly first-hand, peer-reviewed academic literature or reports from major scientific bodies (e.g., IPCC, Nature, WHO). All sources are authoritative, up-to-date, and include a valid, working URL. \\
score 7-8 &  Most key points are properly cited with high-quality, authoritative sources. There may be a few minor statements lacking a direct citation, or a small number of citations may be to high-quality secondary sources (e.g., reputable scientific journalism or a good Wikipedia article) instead of primary literature. All URLs are valid. \\
score 5-6 &  Citations are present but inconsistent; some key claims lack support. The answer relies significantly on secondary sources, news articles, or institutional websites rather than primary academic literature. Some sources may be slightly outdated. \\
score 1-4 &  Citations are largely missing, incorrect, or link to unreliable sources (e.g., blogs, opinion pieces, non-scientific websites). There is little to no use of peer-reviewed literature. Many key claims are unsupported. \\
score 0 &  The answer provides no sources or citations whatsoever, or the sources provided are entirely inappropriate and untrustworthy. \\

\bottomrule
\end{tabular}
\end{table}

\subsubsection{Answers}

\begin{modelanswer}[Model answer (Preferred by Scientists)]{LightGoldenrodYellow}
The optimal data and measurement tools for climate risk modeling and disclosure depend on the specific type of risk being analyzed-physical or transition-and the intended application, such as internal strategy or external reporting. A combination of datasets and analytical tools is typically required rather than a single ``best" solution [2].

\textbf{Data for Climate Risk Modeling}

Effective modeling requires both climate-related and entity-specific data.

\textbf{For Physical Risk Assessment:}
\begin{itemize}[leftmargin=*, nosep]
    \item \textbf{Climate Hazard Data:} Projections of chronic changes (e.g., sea-level rise, mean temperature shifts) and acute events (e.g., hurricanes, floods, wildfires) are fundamental. This data is derived from Global Climate Models (GCMs) and, for higher geographic resolution, dynamically or statistically downscaled Regional Climate Models (RCMs). These models use standardized scenarios, such as the IPCC's Shared Socioeconomic Pathways (SSPs), to project outcomes under different emissions futures [1]. Key data providers include the Copernicus Climate Change Service (C3S) and NASA.

    \item \textbf{Exposure and Vulnerability Data:} High-quality analysis requires granular, asset-level data, including precise geolocations of facilities, infrastructure, and key supply chain points [2]. This is combined with vulnerability data, which describes an asset's susceptibility to a given hazard (e.g., building elevation, flood defense standards) [3].

\end{itemize}

\textbf{For Transition Risk Assessment:} 
\begin{itemize}[leftmargin=*, nosep]
    \item \textbf{Scenario Data:} This includes projections on policy, technology, and market shifts. Authoritative sources provide scenarios detailing potential pathways for carbon prices, energy mixes, and technology adoption rates. The Network for Greening the Financial System (NGFS) and the International Energy Agency (IEA) are key providers of such scenarios [3, 4].

    \item \textbf{Entity-Specific Data:} For both modeling and disclosure, companies must use internal data on their greenhouse gas (GHG) emissions (Scope 1, 2, and 3), energy consumption, revenue dependencies, and strategic plans for decarbonization. This is a core requirement of disclosure standards like IFRS S2 [5].

\end{itemize}

\textbf{Measurement Tools} 
\begin{itemize}[leftmargin=*, nosep]
    \item \textbf{Scenario Analysis:} This is the overarching framework recommended by the Task Force on Climate-related Financial Disclosures (TCFD) for assessing and reporting on climate risks and opportunities. It involves using a set of plausible future scenarios to test the resilience of a business strategy [2].
    \item \textbf{Climate and Economic Models:} Integrated Assessment Models (IAMs) are a key tool for generating transition scenarios, as they link economic activity, energy systems, and the climate [6]. For physical risk, catastrophe (CAT) models, traditionally used in the insurance industry, are adapted to quantify the financial impacts of extreme weather events under climate change [7].
    \item \textbf{Financial Impact Models:} These tools translate the outputs from climate and economic models into financial metrics. This can involve adjusting standard financial models, such as Discounted Cash Flow (DCF) or Value-at-Risk (VaR) models, to incorporate climate-related variables [3].
\end{itemize}

\begin{center}
    \vspace{1ex} 
    \fbox{\textit{The image you are requesting does not exist or is no longer available. [BROKEN LINK]}}

    \vspace{1ex} 
\end{center} 

\textbf{Figure: }The analytical chain for climate risk assessment, showing the progression from scenarios to financial risk impacts. Source: Adapted from NGFS [3].
\vskip 2mm

For disclosure, transparency is paramount. Organizations should report the specific models, scenarios, and key assumptions used in their analysis. It is also critical to acknowledge the inherent uncertainties in long-term climate and economic projections [1, 2].

\textbf{References}

[1]~IPCC, "Climate Change 2021: The Physical Science Basis. Contribution of Working Group I to the Sixth Assessment Report of the Intergovernmental Panel on Climate Change," Cambridge University Press, 2021. [Online]. Available: \url{https://www.ipcc.ch/report/ar6/wg1/}

[2]~Task Force on Climate-related Financial Disclosures, "Recommendations of the Task Force on Climate-related Financial Disclosures," 2017. [Online]. Available: \url{https://www.fsb-tcfd.org/publications/final-recommendations-report/}

[3]~Network for Greening the Financial System, "NGFS Climate Scenarios for central banks and supervisors," 2022. [Online]. Available: \url{https://www.ngfs.net/sites/default/files/medias/documents/ngfs_climate_scenarios_for_central_banks_and_supervisors_phase_iii_september_2022.pdf}

[4]~International Energy Agency, "World Energy Outlook 2023," 2023. [Online]. Available: \url{https://www.iea.org/reports/world-energy-outlook-2023}

[5]~IFRS Foundation, "IFRS S2 Climate-related Disclosures," 2023. [Online]. Available: \url{https://www.ifrs.org/issued-standards/ifrs-sustainability-disclosure-standards/ifrs-s2-climate-related-disclosures/}

[6]~D. van Vuuren et al., "The Shared Socio-economic Pathways: A new framework for climate change research," \textit{Global Environmental Change}, vol. 42, pp. 383--397, 2017. [Online]. Available: \url{https://doi.org/10.1016/j.gloenvcha.2016.10.009}

[7]~S. Hallegatte, "The indirect costs of natural disasters and a review of existing quantitative models," World Bank, Policy Research Working Paper 7375, 2015. [Online]. Available: \url{https://openknowledge.worldbank.org/handle/10986/22465}
\end{modelanswer}

\begin{modelanswer}[Hybrid answer (Human + Gemini 2.5 Flash) (Preferred by Autorater)]{Honeydew}
Climate risk modeling and disclosure are essential for organizations to understand and report the potential financial impacts of climate change [1]. This involves assessing both physical risks, such as extreme weather events and sea-level rise, and transition risks, including policy changes, technological advancements, and market shifts [1]. Physical risk is fundamentally a combination of climate hazard, exposure, and vulnerability [2].

Effective climate risk modeling necessitates the use of diverse and high-quality data. For physical risks, crucial data types include climate science data (projections from climate models), geographic information, historical loss data [1, 2], observations, reanalyses, and climate indices like the Standardized Precipitation Index (SPI) or Standardized Precipitation Evapotranspiration Index (SPEI) [2]. High-resolution and even hyper-local data are increasingly important for assessing localized impacts [6, 2]. For transition risks, data encompasses information on policies, technology trends, market dynamics, and company-specific operational data such as energy use and emissions [1]. The quality, granularity, and forward-looking nature of data are critical for robust modeling [1, 3, 2]. Relevant data sources include outputs from climate models, remote sensing, ground observations [2], data from organizations like CDP [7], and company financial filings (e.g., 10-K reports) [5].

Measurement tools employed in climate risk modeling span a wide range of approaches. These include statistical models, scenario analysis, and stress testing [1, 3]. More complex tools involve integrated climate and economic models [1]. To translate large-scale climate data to local impacts, downscaling techniques, both dynamical and statistical, are frequently used [2, 6]. Textual analysis of corporate disclosures is also utilized to assess reported risks [5]. A conceptual framework can help outline the necessary data and appropriate risk measurement approaches (Figure 1) [4, 3].

\begin{center}
    \vspace{1ex}
    \fbox{\textit{Image: Climate Risk Assessment Framework for Banks and Supervisors} [\href{https://analystprep.com/study-notes/wp-content/uploads/2023/03/A-conceptual-climate-risk-assessment-framework-for-banks-and-supervisors.jpg}{ImageLink}]}
    \vspace{1ex}
\end{center} 

\textbf{Figure:} A conceptual framework for climate risk assessment outlines key steps, including identifying data needs and applying risk measurement approaches [4].
\vskip 2mm

Significant challenges exist in climate risk modeling and disclosure. These include issues with data availability, quality, consistency, and scalability [1, 3, 2]. Modeling inherently involves uncertainties, particularly over longer time horizons [1, 2, 3]. Furthermore, there are challenges related to the transparency and comparability of the tools and the resulting disclosures [3]. Many tools lack public transparency regarding their setup and underlying assumptions, and there is a need for improved communication of output uncertainties and limitations [3]. The lack of peer-reviewed approaches for some tools also poses a challenge [3].

Ultimately, there is no single "best" type of data or measurement tool. Effective climate risk modeling and disclosure require integrating diverse data sources and applying appropriate tools tailored to the specific risks being assessed, the sector, and relevant disclosure standards [1, 3]. Forward-looking assessments and scenario analysis remain key components of a robust approach [1, 2, 3].

\textbf{References}

[1]~PwC, "Climate risk modeling: What you need to know to get started," PwC, 2021. Available: \url{https://www.pwc.com/us/en/services/esg/library/climate-risk-modeling.html}

[2]~A. Tantet and P. Tankov, "Climate Data for Physical Risk Assessment in Finance," SSRN, 2019. Available: \url{https://ssrn.com/abstract=3480156}

[3]~J. A. Bingler and C. Colesanti Senni, "Taming the Green Swan: a criteria-based analysis to improve the understanding of climate-related financial risk assessment tools," \textit{Climate Policy}, vol. 22, no. 3, pp. 356--370, 2022. Available: \url{https://doi.org/10.1080/14693062.2022.2032569}

[4]~CFA Institute, "Climate-Related Financial Risks – Measurement Methodologies," CFA Institute, 2023. Available: 
\href{https://analystprep.com/study-notes/wp-content/uploads/2023/03/A-conceptual-climate-risk-assessment-framework-for-banks-and-supervisors.jpg}{\texttt{analystprep.com ... assessment-framework-for-banks-and-supervisors.jpg}}

[5]~H. Berkman, J. Jona, and N. Soderstrom, "Firm-specific climate risk and market valuation," \textit{Accounting, Organizations and Society}, vol. 112, 2024. Available: \url{file:///uploaded/1-s2.0-S0361368224000072-main.pdf}

[6]~Cotality, "Comprehensive Climate Risk Analytics," Cotality. Available: \url{https://www.cotality.com/products/climate-risk-analytics}

[7]~E. Ilhan, P. Krueger, Z. Sautner, and L. T. Starks, "Climate Risk Disclosure and Institutional Investors," SSRN, 2022. Available: \url{file:///uploaded/ssrn-3437178.pdf}
\end{modelanswer}

\subsubsection{Verdicts and Autorater Rationale}

\paragraph{Scientists Verdict:} Model answer wins over Hybrid.\\
Dimensions favoring model answer:
\begin{itemize}
    \item presentation\_style
    \item presentation\_clarity
\end{itemize}

\paragraph{Autorater Verdict:} Hybrid wins over model answer.\\
Dimensions favoring model answer: 
\begin{itemize}
    \item knowledge\_specificity.
\end{itemize}

Dimensions favoring Hybrid: 
\begin{itemize}
    \item knowledge\_uncertainty
    \item knowledge\_accuracy
    \item knowledge\_completeness
    \item citations\_source\_quality
\end{itemize}

\paragraph{Autorater Explanation (Excerpt)}
\begin{quote}
"Answer 2 is preferred because it offers a more comprehensive and nuanced understanding of climate risk modeling and disclosure. While Answer 1 (model answer) provides excellent specificity regarding certain data types and models, \textbf{Answer 2 significantly surpasses it in its detailed acknowledgment and discussion of the inherent challenges, uncertainties, and limitations within this complex field.} This crucial aspect is vital for a complete and realistic assessment of climate risk. Additionally, Answer 2 demonstrates better source quality... Answer 1's reliance on invalid references for key frameworks like NGFS and IFRS S2 weakens the verifiability of important details..."
\end{quote}
Note that above Answer 1 refers to the model answer and Answer 2 refers to the Hybrid answer.

\paragraph{Key Rubric Criterion: Nuance and Acknowledgment of Challenges (Excerpt)}
\begin{itemize}
    \item \textbf{Model Rating (5.0/10):} ``Answer 1 briefly acknowledges 'inherent uncertainties...' in its concluding paragraph, but does not elaborate further."
    \item \textbf{Hybrid Rating (9.5/10):} ``Answer 2 dedicates a significant section to 'Significant challenges,' discussing data availability, quality, consistency, scalability, inherent uncertainties (long time horizons), transparency, comparability, and the lack of peer-reviewed approaches. This provides a sophisticated and detailed discussion."
\end{itemize}

\paragraph{Supplemental Materials Check (Excerpt)}
\begin{itemize}
    \item The reference URL for IFRS S2 (\textit{Used by model answer, Ref [5]}) is NOT valid.
    \item The reference URL for NGFS (\textit{Used by model answer, Ref [3]}) is NOT valid.
    \item Answer 1 (\textit{model answer}) contains 7 references. 2 of the 7 references NOT valid.
    \item Answer 2 (\textit{Hybrid}) contains 7 references. 1 of the 7 references NOT valid (Ref [2], excluding local PDF links).
\end{itemize}
In the items above, the information regarding model identities in \textit{italics} has been added for readability. The source for any answer is never available to the autorater.

%% file: main.bbl
\begin{thebibliography}{72}
\providecommand{\natexlab}[1]{#1}
\providecommand{\url}[1]{\texttt{#1}}
\expandafter\ifx\csname urlstyle\endcsname\relax
  \providecommand{\doi}[1]{doi: #1}\else
  \providecommand{\doi}{doi: \begingroup \urlstyle{rm}\Url}\fi

\bibitem[Arora et~al.(2025)Arora, Wei, Hicks, Bowman, Qui{\~n}onero-Candela,
  Tsimpourlas, Sharman, Shah, Vallone, Beutel, et~al.]{arora2025healthbench}
R.~K. Arora, J.~Wei, R.~S. Hicks, P.~Bowman, J.~Qui{\~n}onero-Candela,
  F.~Tsimpourlas, M.~Sharman, M.~Shah, A.~Vallone, A.~Beutel, et~al.
\newblock Healthbench: Evaluating large language models towards improved human
  health.
\newblock \emph{arXiv preprint arXiv:2505.08775}, 2025.

\bibitem[Asai et~al.(2024)Asai, Wu, Wang, Sil, and Hajishirzi]{asai2024selfrag}
A.~Asai, Z.~Wu, Y.~Wang, A.~Sil, and H.~Hajishirzi.
\newblock Self-rag: Learning to retrieve, generate, and critique through
  self-reflection.
\newblock In \emph{The Twelfth International Conference on Learning
  Representations (ICLR)}, 2024.

\bibitem[Bajpai et~al.(2024)Bajpai, Chatterjee, Dutta, and
  Chakraborty]{Bajpai2024CanLRA}
P.~Bajpai, N.~Chatterjee, S.~Dutta, and T.~Chakraborty.
\newblock Can llms replace neil degrasse tyson? evaluating the reliability of
  llms as science communicators.
\newblock \emph{ArXiv}, abs/2409.14037, 2024.
\newblock URL \url{https://api.semanticscholar.org/CorpusId:272826664}.

\bibitem[Bowman et~al.(2022)Bowman, Hyun, Durmus, Askell, Kernion, Hernandez,
  Hubinger, Kernion, Lovitt, DasSarma, Brown, Olsson, Amodei, Kaplan,
  McCandlish, and Henighan]{bowman2022measuring}
S.~R. Bowman, J.~Hyun, E.~Durmus, A.~Askell, A.~Kernion, D.~Hernandez,
  E.~Hubinger, J.~Kernion, L.~Lovitt, N.~DasSarma, T.~B. Brown, C.~Olsson,
  D.~Amodei, J.~Kaplan, S.~McCandlish, and T.~Henighan.
\newblock Measuring progress on scalable oversight for large language models,
  2022.

\bibitem[Bran et~al.(2023)Bran, Cox, Schilter, Baldassari, White, and
  Schwaller]{Bran2023AugmentingLLA}
A.~M. Bran, S.~Cox, O.~Schilter, C.~Baldassari, A.~D. White, and P.~Schwaller.
\newblock Augmenting large language models with chemistry tools.
\newblock \emph{Nature Machine Intelligence}, 6:\penalty0 525 -- 535, 2023.
\newblock URL \url{https://www.ncbi.nlm.nih.gov/pubmed/38799228}.

\bibitem[Bulian et~al.(2024)Bulian, Schäfer, Amini, Lam, Ciaramita, Gaiarin,
  H\"{u}bscher, Buck, Mede, Leippold, and
  Strau{\ss}]{assessinglargelanguagemodels}
J.~Bulian, M.~S. Schäfer, A.~Amini, H.~Lam, M.~Ciaramita, B.~Gaiarin, M.~C.
  H\"{u}bscher, C.~Buck, N.~G. Mede, M.~Leippold, and N.~Strau{\ss}.
\newblock Assessing large language models on climate information.
\newblock In \emph{Proceedings of the 41st International Conference on Machine
  Learning}, 2024.

\bibitem[Byun et~al.(2024)Byun, Vasicek, and Seppi]{byun2024reference}
C.~Byun, P.~Vasicek, and K.~Seppi.
\newblock This reference does not exist: an exploration of llm citation
  accuracy and relevance.
\newblock In \emph{Proceedings of the Third Workshop on Bridging
  Human--Computer Interaction and Natural Language Processing}, pages 28--39,
  2024.

\bibitem[Calderon et~al.(2025)Calderon, Reichart, and Dror]{Calderon2025TheAAA}
N.~Calderon, R.~Reichart, and R.~Dror.
\newblock The alternative annotator test for llm-as-a-judge: How to
  statistically justify replacing human annotators with llms.
\newblock \emph{arXiv preprint arXiv:2501.10970}, 2025.

\bibitem[Chiang et~al.(2024{\natexlab{a}})Chiang, Zheng, Sheng, Angelopoulos,
  Li, Li, Zhang, Zhu, Jordan, Gonzalez, and
  Stoica]{chiang2024chatbotarenaopenplatform}
W.-L. Chiang, L.~Zheng, Y.~Sheng, A.~N. Angelopoulos, T.~Li, D.~Li, H.~Zhang,
  B.~Zhu, M.~Jordan, J.~E. Gonzalez, and I.~Stoica.
\newblock Chatbot arena: An open platform for evaluating llms by human
  preference, 2024{\natexlab{a}}.
\newblock URL \url{https://arxiv.org/abs/2403.04132}.

\bibitem[Chiang et~al.(2024{\natexlab{b}})Chiang, Zheng, Sheng, Angelopoulos,
  Li, Li, Zhu, Zhang, Jordan, Gonzalez, and Stoica]{chatbotareana}
W.-L. Chiang, L.~Zheng, Y.~Sheng, A.~N. Angelopoulos, T.~Li, D.~Li, B.~Zhu,
  H.~Zhang, M.~I. Jordan, J.~E. Gonzalez, and I.~Stoica.
\newblock Chatbot arena: an open platform for evaluating llms by human
  preference.
\newblock In \emph{Proceedings of the 41st International Conference on Machine
  Learning}, ICML'24, 2024{\natexlab{b}}.

\bibitem[Dinh et~al.(2024)Dinh, Mullov, Barmann, Li, Liu, Rei{\ss}, Lee,
  Lerzer, Ternava, Gao, Waibel, Asfour, Beigl, Stiefelhagen, Dachsbacher, Bohm,
  and Niehues]{Dinh2024SciExBLA}
T.~A. Dinh, C.~Mullov, L.~Barmann, Z.~Li, D.~Liu, S.~Rei{\ss}, J.~Lee,
  N.~Lerzer, F.~Ternava, J.~Gao, A.~Waibel, T.~Asfour, M.~Beigl,
  R.~Stiefelhagen, C.~Dachsbacher, K.~Bohm, and J.~Niehues.
\newblock Sciex: Benchmarking large language models on scientific exams with
  human expert grading and automatic grading.
\newblock In \emph{Conference on Empirical Methods in Natural Language
  Processing}, 2024.
\newblock URL \url{https://api.semanticscholar.org/CorpusId:270559604}.

\bibitem[D{'}Souza et~al.(2025)D{'}Souza, Babaei~Giglou, and
  M{\"u}nch]{dsouza-etal-2025-yescieval}
J.~D{'}Souza, H.~Babaei~Giglou, and Q.~M{\"u}nch.
\newblock {YES}ci{E}val: Robust {LLM}-as-a-judge for scientific question
  answering.
\newblock In W.~Che, J.~Nabende, E.~Shutova, and M.~T. Pilehvar, editors,
  \emph{Proceedings of the 63rd Annual Meeting of the Association for
  Computational Linguistics (Volume 1: Long Papers)}, pages 13749--13783,
  Vienna, Austria, July 2025. Association for Computational Linguistics.
\newblock ISBN 979-8-89176-251-0.
\newblock \doi{10.18653/v1/2025.acl-long.675}.
\newblock URL \url{https://aclanthology.org/2025.acl-long.675/}.

\bibitem[Eger et~al.(2025)Eger, Cao, D'Souza, Geiger, Greisinger, Gross, Hou,
  Krenn, Lauscher, Li, et~al.]{eger2025transforming}
S.~Eger, Y.~Cao, J.~D'Souza, A.~Geiger, C.~Greisinger, S.~Gross, Y.~Hou,
  B.~Krenn, A.~Lauscher, Y.~Li, et~al.
\newblock Transforming science with large language models: A survey on
  {AI-}assisted scientific discovery, experimentation, content generation, and
  evaluation.
\newblock \emph{CoRR}, 2025.

\bibitem[Fan et~al.(2019)Fan, Jernite, Perez, Scott, de~Vries, and
  Kiela]{fan2019eli5}
A.~Fan, Y.~Jernite, E.~Perez, D.~Scott, H.~de~Vries, and D.~Kiela.
\newblock {ELI5}: Long form question answering.
\newblock In \emph{Proceedings of the 57th Annual Meeting of the Association
  for Computational Linguistics}, pages 3558--3567, 2019.

\bibitem[Gao et~al.(2023)Gao, Yen, Yu, and Chen]{gao2023enabling}
T.~Gao, H.~Yen, J.~Yu, and D.~Chen.
\newblock Enabling large language models to generate text with citations.
\newblock \emph{arXiv preprint arXiv:2305.14627}, 2023.

\bibitem[Gera et~al.(2025)Gera, Boni, Perlitz, Bar-Haim, Edelstein, and
  Yehudai]{gera2025justrank}
A.~Gera, O.~Boni, Y.~Perlitz, R.~Bar-Haim, L.~Edelstein, and A.~Yehudai.
\newblock Justrank: Benchmarking llm judges for system ranking.
\newblock In \emph{Annual Meeting of the Association for Computational
  Linguistics}, 2025.

\bibitem[Gu et~al.(2024)Gu, Jiang, Shi, Tan, Zhai, Xu, Li, Shen, Ma, Liu,
  et~al.]{gu2024survey}
J.~Gu, X.~Jiang, Z.~Shi, H.~Tan, X.~Zhai, C.~Xu, W.~Li, Y.~Shen, S.~Ma, H.~Liu,
  et~al.
\newblock A survey on {LLM}-as-a-judge.
\newblock \emph{arXiv preprint arXiv:2411.15594}, 2024.

\bibitem[Gudibande et~al.(2024)Gudibande, Wallace, Snell, Geng, Liu, Abbeel,
  Levine, and Song]{gudibande2024false}
A.~Gudibande, E.~Wallace, C.~Snell, X.~Geng, H.~Liu, P.~Abbeel, S.~Levine, and
  D.~Song.
\newblock The false promise of imitating proprietary llms.
\newblock In \emph{The Twelfth International Conference on Learning
  Representations (ICLR)}, 2024.
\newblock URL \url{https://openreview.net/forum?id=u3lyx01O7v}.

\bibitem[Hashemi et~al.(2024)Hashemi, Eisner, Rosset, Van~Durme, and
  Kedzie]{hashemi-etal-2024-llm}
H.~Hashemi, J.~Eisner, C.~Rosset, B.~Van~Durme, and C.~Kedzie.
\newblock {LLM}-rubric: A multidimensional, calibrated approach to automated
  evaluation of natural language texts.
\newblock In L.-W. Ku, A.~Martins, and V.~Srikumar, editors, \emph{Proceedings
  of the 62nd Annual Meeting of the Association for Computational Linguistics
  (Volume 1: Long Papers)}, pages 13806--13834, Bangkok, Thailand, Aug. 2024.
  Association for Computational Linguistics.
\newblock \doi{10.18653/v1/2024.acl-long.745}.
\newblock URL \url{https://aclanthology.org/2024.acl-long.745/}.

\bibitem[He et~al.(2024)He, Luo, Bai, Hu, Thai, Shen, Hu, Han, Huang, Zhang,
  et~al.]{he2024olympiadbench}
C.~He, R.~Luo, Y.~Bai, S.~Hu, Z.~L. Thai, J.~Shen, J.~Hu, X.~Han, Y.~Huang,
  Y.~Zhang, et~al.
\newblock Olympiadbench: A challenging benchmark for promoting agi with
  olympiad-level bilingual multimodal scientific problems.
\newblock \emph{arXiv preprint arXiv:2402.14008}, 2024.

\bibitem[Hendrycks et~al.(2020)Hendrycks, Burns, Basart, Zou, Mazeika, Song,
  and Steinhardt]{hendrycks2020measuring}
D.~Hendrycks, C.~Burns, S.~Basart, A.~Zou, M.~Mazeika, D.~Song, and
  J.~Steinhardt.
\newblock Measuring massive multitask language understanding.
\newblock \emph{arXiv preprint arXiv:2009.03300}, 2020.

\bibitem[Jeong et~al.(2024)Jeong, Park, Hong, Lee, and
  Choo]{jeong2024comparative}
H.~Jeong, C.~Park, J.~Hong, H.~Lee, and J.~Choo.
\newblock The comparative trap: Pairwise comparisons amplifies biased
  preferences of llm evaluators.
\newblock \emph{arXiv preprint arXiv:2406.12319}, 2024.

\bibitem[Ji et~al.(2024)Ji, Liu, Du, and Ng]{ji2024chain}
B.~Ji, H.~Liu, M.~Du, and S.-K. Ng.
\newblock Chain-of-thought improves text generation with citations in large
  language models.
\newblock \emph{Proceedings of the AAAI Conference on Artificial Intelligence},
  38\penalty0 (16):\penalty0 18345--18353, 2024.

\bibitem[Ji et~al.(2023)Ji, Lee, Frieske, Yu, Su, Xu, Ishii, Bang, Madotto, and
  Fung]{ji2023survey}
Z.~Ji, N.~Lee, R.~Frieske, T.~Yu, D.~Su, Y.~Xu, E.~Ishii, Y.~Bang, A.~Madotto,
  and P.~Fung.
\newblock Survey of hallucination in natural language generation.
\newblock \emph{ACM Computing Surveys}, 55\penalty0 (12):\penalty0 1--38, 2023.

\bibitem[Justen(2025)]{justen2025llmsoutperformexpertschallenging}
L.~Justen.
\newblock Llms outperform experts on challenging biology benchmarks.
\newblock \emph{ArXiv}, 2025.
\newblock URL \url{https://arxiv.org/abs/2505.06108}.

\bibitem[Kaack et~al.(2022)Kaack, Donti, Strubell, Genc, and
  Rolnick]{kaack2022aligning}
L.~H. Kaack, P.~L. Donti, E.~Strubell, G.~Genc, and D.~Rolnick.
\newblock Aligning artificial intelligence with climate change mitigation.
\newblock \emph{Nature Climate Change}, 12\penalty0 (6):\penalty0 518--528,
  2022.

\bibitem[Kiela et~al.(2021)Kiela, Bartolo, Kadian, R{\'e}tornaz, K{\"o}pf,
  Singh, Wang, Chen, Parikh, et~al.]{kiela2021dynabench}
D.~Kiela, M.~Bartolo, A.~Kadian, T.~R{\'e}tornaz, J.~K{\"o}pf, A.~Singh,
  S.~Wang, G.~Chen, D.~Parikh, et~al.
\newblock Dynabench: Rethinking benchmarking in nlp.
\newblock In \emph{Proceedings of the 2021 Conference of the North American
  Chapter of the Association for Computational Linguistics: Human Language
  Technologies (NAACL-HLT)}, pages 4110--4124, 2021.

\bibitem[Kim et~al.(2024)Kim, Shin, Cho, Jang, Longpre, Lee, Lee, Kim, Thorne,
  and Seo]{kim2024prometheus}
S.~Kim, J.~Shin, Y.~Cho, J.~Jang, S.~Longpre, H.~Lee, S.-W. Lee, S.~Kim,
  J.~Thorne, and M.~Seo.
\newblock Prometheus: Inducing fine-grained evaluation capability in lm
  evaluators.
\newblock In \emph{The Twelfth International Conference on Learning
  Representations (ICLR)}, 2024.
\newblock URL \url{https://openreview.net/forum?id=T54g8j0g1n}.

\bibitem[Laurent et~al.(2024)Laurent, Janizek, Ruzo, Hinks, Hammerling,
  Narayanan, Ponnapati, White, and Rodriques]{laurent2024lab}
J.~M. Laurent, J.~D. Janizek, M.~Ruzo, M.~M. Hinks, M.~J. Hammerling,
  S.~Narayanan, M.~Ponnapati, A.~D. White, and S.~G. Rodriques.
\newblock Lab-bench: Measuring capabilities of language models for biology
  research.
\newblock \emph{arXiv preprint arXiv:2407.10362}, 2024.

\bibitem[Laurito et~al.(2025)Laurito, Davis, Grietzer, Gaven{\v{c}}iak,
  B{\"o}hm, and Kulveit]{laurito2025aiai}
W.~Laurito, B.~Davis, P.~Grietzer, T.~Gaven{\v{c}}iak, A.~B{\"o}hm, and
  J.~Kulveit.
\newblock Ai-ai bias: Large language models favor communications generated by
  large language models.
\newblock \emph{Proceedings of the National Academy of Sciences}, 122\penalty0
  (31):\penalty0 e2415697122, 2025.

\bibitem[Lewis et~al.(2020)Lewis, Perez, Piktus, Petroni, Karpukhin, Goyal,
  K{\"u}ttler, Lewis, Yih, Rockt{\"a}schel, et~al.]{lewis2020retrieval}
P.~Lewis, E.~Perez, A.~Piktus, F.~Petroni, V.~Karpukhin, N.~Goyal,
  H.~K{\"u}ttler, M.~Lewis, W.-t. Yih, T.~Rockt{\"a}schel, et~al.
\newblock Retrieval-augmented generation for knowledge-intensive nlp tasks.
\newblock \emph{Advances in Neural Information Processing Systems},
  33:\penalty0 9459--9474, 2020.

\bibitem[Li et~al.(2025)Li, Deng, Lu, and
  Yuan]{li2025atmosscibenchevaluatingrecentadvance}
C.~Li, W.~Deng, M.~Lu, and B.~Yuan.
\newblock Atmossci-bench: Evaluating the recent advance of large language model
  for atmospheric science, 2025.
\newblock URL \url{https://arxiv.org/abs/2502.01159}.

\bibitem[Li et~al.(2024{\natexlab{a}})Li, Sun, Hu, Liu, Hu, Liu, and
  Zhang]{DBLP:conf/acl/LiSHLH0Z24}
D.~Li, Z.~Sun, B.~Hu, Z.~Liu, X.~Hu, X.~Liu, and M.~Zhang.
\newblock Improving attributed text generation of large language models via
  preference learning.
\newblock In L.~Ku, A.~Martins, and V.~Srikumar, editors, \emph{Findings of the
  Association for Computational Linguistics, {ACL} 2024, Bangkok, Thailand and
  virtual meeting, August 11-16, 2024}, pages 5079--5101. Association for
  Computational Linguistics, 2024{\natexlab{a}}.
\newblock URL \url{https://aclanthology.org/2024.findings-acl.301}.

\bibitem[Li et~al.(2024{\natexlab{b}})Li, Dong, Chen, Su, Zhou, Ai, Ye, and
  Liu]{li2024llms}
H.~Li, Q.~Dong, J.~Chen, H.~Su, Y.~Zhou, Q.~Ai, Z.~Ye, and Y.~Liu.
\newblock {LLMs}-as-judges: a comprehensive survey on llm-based evaluation
  methods.
\newblock \emph{arXiv preprint arXiv:2412.05579}, 2024{\natexlab{b}}.

\bibitem[Liu et~al.(2025)Liu, Han, Luo, Wang, Lu, Zhu, Wang, Li, Chen, Qu, Liu,
  Cui, Yaish, Chen, Hao, Li, Li, Cohan, Xu, Gerstein, Zou, and
  Zhao]{Liu2025TowardsAIE}
T.~Liu, S.~Han, X.~Luo, H.~Wang, P.~Lu, B.~Zhu, Y.~Wang, K.~Li, J.~Chen, R.~Qu,
  Y.~Liu, X.~Cui, A.~Yaish, Y.~Chen, M.~Hao, C.~Li, K.~Li, A.~Cohan, H.~Xu,
  M.~Gerstein, J.~Zou, and H.~Zhao.
\newblock Towards artificial intelligence research assistant for
  expert-involved learning.
\newblock \emph{ArXiv}, abs/2505.04638, 2025.
\newblock URL \url{https://api.semanticscholar.org/CorpusId:278394723}.

\bibitem[{Long, et al.}(2025)]{hle}
P.~{Long, et al.}
\newblock Humanity's {L}ast {E}xam, 2025.
\newblock URL \url{https://arxiv.org/abs/2501.14249}.

\bibitem[Ma et~al.(2024)Ma, Gou, Hao, Xu, Wang, Pan, Yang, Cao, and
  Sun]{ma-etal-2024-sciagent}
Y.~Ma, Z.~Gou, J.~Hao, R.~Xu, S.~Wang, L.~Pan, Y.~Yang, Y.~Cao, and A.~Sun.
\newblock {S}ci{A}gent: Tool-augmented language models for scientific
  reasoning.
\newblock In Y.~Al-Onaizan, M.~Bansal, and Y.-N. Chen, editors,
  \emph{Proceedings of the 2024 Conference on Empirical Methods in Natural
  Language Processing}, pages 15701--15736, Miami, Florida, USA, Nov. 2024.
  Association for Computational Linguistics.
\newblock \doi{10.18653/v1/2024.emnlp-main.880}.
\newblock URL \url{https://aclanthology.org/2024.emnlp-main.880/}.

\bibitem[Malikussaid and Nuha(2025)]{Malikussaid2025BridgingTPD}
Malikussaid and H.~Nuha.
\newblock Bridging the plausibility-validity gap by fine-tuning a
  reasoning-enhanced llm for chemical synthesis and discovery.
\newblock \emph{ArXiv}, abs/2507.07328, 2025.
\newblock URL \url{https://api.semanticscholar.org/CorpusId:280280955}.

\bibitem[Manivannan et~al.(2024)Manivannan, Jafari, Eranky, Ho, Yu,
  Watson-Parris, Ma, Bergen, and Berg-Kirkpatrick]{manivannan2024climaqa}
V.~V. Manivannan, Y.~Jafari, S.~Eranky, S.~Ho, R.~Yu, D.~Watson-Parris, Y.~Ma,
  L.~Bergen, and T.~Berg-Kirkpatrick.
\newblock Climaqa: An automated evaluation framework for climate foundation
  models.
\newblock \emph{arXiv preprint arXiv:2410.16701}, 2024.

\bibitem[Menick et~al.(2022)Menick, Trebacz, Mikulik, Aslanides, Song,
  Chadwick, Glaese, Young, Campbell-Gillingham, Irving, Uesato, Rae, Shi, and
  Huang]{menick2022teaching}
J.~Menick, M.~Trebacz, V.~Mikulik, J.~Aslanides, F.~Song, M.~Chadwick,
  M.~Glaese, S.~Young, L.~Campbell-Gillingham, G.~Irving, J.~Uesato, J.~W. Rae,
  K.~Shi, and S.~Huang.
\newblock Teaching language models to support answers with verified quotes,
  2022.

\bibitem[Min et~al.(2023)Min, Krishna, Lyu, Lewis, Yih, Koh, Gururangan, Choi,
  Hajishirzi, and Zettlemoyer]{min2023factscore}
S.~Min, K.~Krishna, X.~Lyu, M.~Lewis, W.-t. Yih, P.~W. Koh, S.~Gururangan,
  Y.~Choi, H.~Hajishirzi, and L.~Zettlemoyer.
\newblock {FActScore}: Fine-grained atomic evaluation of factual precision in
  long form text generation.
\newblock In \emph{Proceedings of the 2023 Conference on Empirical Methods in
  Natural Language Processing (EMNLP)}, 2023.

\bibitem[Nakano et~al.(2021)Nakano, Hilton, Balaji, Wu, Ouyang, Kim, Hesse,
  Jain, Kosaraju, Saunders, Jiang, Cobbe, Eloundou, Krueger, Button, Knight,
  Chess, and Schulman]{nakano2021webgpt}
R.~Nakano, J.~Hilton, S.~Balaji, J.~Wu, L.~Ouyang, C.~Kim, C.~Hesse, S.~Jain,
  V.~Kosaraju, W.~Saunders, X.~Jiang, K.~Cobbe, T.~Eloundou, G.~Krueger,
  K.~Button, M.~Knight, B.~Chess, and J.~Schulman.
\newblock Webgpt: Browser-assisted question-answering with human feedback,
  2021.

\bibitem[Noy and Zhang(2023)]{noy2023experimental}
S.~Noy and W.~Zhang.
\newblock Experimental evidence on the productivity effects of generative
  artificial intelligence.
\newblock \emph{Science}, 381\penalty0 (6654):\penalty0 187--192, 2023.

\bibitem[Oh et~al.(2024)Oh, Kim, Cha, and Oh]{oh-etal-2024-generative}
J.~Oh, E.~Kim, I.~Cha, and A.~Oh.
\newblock The generative {AI} paradox in evaluation: ``what it can solve, it
  may not evaluate''.
\newblock In N.~Falk, S.~Papi, and M.~Zhang, editors, \emph{Proceedings of the
  18th Conference of the European Chapter of the Association for Computational
  Linguistics: Student Research Workshop}, pages 248--257, St. Julian{'}s,
  Malta, Mar. 2024. Association for Computational Linguistics.
\newblock \doi{10.18653/v1/2024.eacl-srw.19}.
\newblock URL \url{https://aclanthology.org/2024.eacl-srw.19/}.

\bibitem[Page et~al.(1998)Page, Brin, Motwani, and Winograd]{Page1998PageRank}
L.~Page, S.~Brin, R.~Motwani, and T.~Winograd.
\newblock {The PageRank Citation Ranking: Bringing Order to the Web}.
\newblock Technical report, Stanford Digital Library Technologies Project,
  1998.
\newblock URL
  \url{http://citeseerx.ist.psu.edu/viewdoc/summary?doi=10.1.1.31.1768}.

\bibitem[Panickssery et~al.(2024)Panickssery, Bowman, and
  Feng]{panickssery2024llm}
A.~Panickssery, S.~Bowman, and S.~Feng.
\newblock {LLM} evaluators recognize and favor their own generations.
\newblock \emph{Advances in Neural Information Processing Systems},
  37:\penalty0 68772--68802, 2024.

\bibitem[Rein et~al.(2024)Rein, Hou, Stickland, Petty, Pang, Dirani, Michael,
  and Bowman]{gpqa}
D.~Rein, B.~L. Hou, A.~C. Stickland, J.~Petty, R.~Y. Pang, J.~Dirani,
  J.~Michael, and S.~R. Bowman.
\newblock {GPQA}: A graduate-level google-proof q\&a benchmark.
\newblock In \emph{First Conference on Language Modeling}, 2024.
\newblock URL \url{https://openreview.net/forum?id=Ti67584b98}.

\bibitem[Roberts et~al.(2024)Roberts, Han, Houlsby, and
  Albanie]{roberts2024scifibench}
J.~Roberts, K.~Han, N.~Houlsby, and S.~Albanie.
\newblock Scifibench: Benchmarking large multimodal models for scientific
  figure interpretation.
\newblock \emph{Advances in Neural Information Processing Systems},
  37:\penalty0 18695--18728, 2024.

\bibitem[Robertson and Koyejo(2025)]{robertson2025let}
Z.~Robertson and S.~Koyejo.
\newblock Let's measure information step-by-step: Llm-based evaluation beyond
  vibes.
\newblock \emph{arXiv preprint arXiv:2508.05469}, 2025.

\bibitem[Rolnick et~al.(2022)Rolnick, Donti, Kaack, Kochanski, Lacoste,
  Sankaran, Ross, Milojevic-Dupont, Jaques, Waldman-Brown,
  et~al.]{rolnick2022tackling}
D.~Rolnick, P.~L. Donti, L.~H. Kaack, K.~Kochanski, A.~Lacoste, K.~Sankaran,
  A.~S. Ross, N.~Milojevic-Dupont, N.~Jaques, A.~Waldman-Brown, et~al.
\newblock Tackling climate change with machine learning.
\newblock \emph{ACM Computing Surveys (CSUR)}, 55\penalty0 (2):\penalty0 1--96,
  2022.

\bibitem[Ruan et~al.(2025)Ruan, Nair, Cao, Liu, Munir, Pollens-Dempsey, Chiang,
  Kates, David, Chen, et~al.]{ruan2025expertlongbench}
J.~Ruan, I.~Nair, S.~Cao, A.~Liu, S.~Munir, M.~Pollens-Dempsey, T.~Chiang,
  L.~Kates, N.~David, S.~Chen, et~al.
\newblock Expertlongbench: Benchmarking language models on expert-level
  long-form generation tasks with structured checklists.
\newblock \emph{arXiv preprint arXiv:2506.01241}, 2025.

\bibitem[Russell(2019)]{Russell2019HumanCompatible}
S.~J. Russell.
\newblock \emph{Human Compatible: Artificial Intelligence and the Problem of
  Control}.
\newblock Viking / Allen Lane, 2019.

\bibitem[Shen et~al.(2025)Shen, Zhou, Chen, Qiu, Liu, Liu, and
  Zhao]{shen-etal-2025-transparentize}
J.~Shen, T.~Zhou, Y.~Chen, D.~Qiu, S.~Liu, K.~Liu, and J.~Zhao.
\newblock Transparentize the internal and external knowledge utilization in
  {LLM}s with trustworthy citation.
\newblock In W.~Che, J.~Nabende, E.~Shutova, and M.~T. Pilehvar, editors,
  \emph{Findings of the Association for Computational Linguistics: ACL 2025},
  pages 17858--17877, Vienna, Austria, July 2025. Association for Computational
  Linguistics.
\newblock ISBN 979-8-89176-256-5.
\newblock \doi{10.18653/v1/2025.findings-acl.919}.
\newblock URL \url{https://aclanthology.org/2025.findings-acl.919/}.

\bibitem[Stelmakh et~al.(2022)Stelmakh, Luan, Dhingra, and
  Chang]{stelmakh2022asqa}
I.~Stelmakh, Y.~Luan, B.~Dhingra, and M.-W. Chang.
\newblock {ASQA}: Factoid questions meet long-form answers.
\newblock In Y.~Goldberg, Z.~Kozareva, and Y.~Zhang, editors, \emph{Proceedings
  of the 2022 Conference on Empirical Methods in Natural Language Processing},
  pages 8273--8288, Abu Dhabi, United Arab Emirates, Dec. 2022. Association for
  Computational Linguistics.
\newblock \doi{10.18653/v1/2022.emnlp-main.566}.
\newblock URL \url{https://aclanthology.org/2022.emnlp-main.566/}.

\bibitem[Sun et~al.(2024)Sun, Li, Yuan, Yuan, Li, and
  Liu]{sun-etal-2024-critique}
S.~Sun, J.~Li, W.~Yuan, R.~Yuan, W.~Li, and P.~Liu.
\newblock The critique of critique.
\newblock In L.-W. Ku, A.~Martins, and V.~Srikumar, editors, \emph{Findings of
  the Association for Computational Linguistics: ACL 2024}, pages 9077--9096,
  Bangkok, Thailand, Aug. 2024. Association for Computational Linguistics.
\newblock \doi{10.18653/v1/2024.findings-acl.538}.
\newblock URL \url{https://aclanthology.org/2024.findings-acl.538/}.

\bibitem[Tan et~al.(2025)Tan, Zhuang, Montgomery, Tang, Cuadron, Wang, Popa,
  and Stoica]{tan2025judgebench}
S.~Tan, S.~Zhuang, K.~Montgomery, W.~Y. Tang, A.~Cuadron, C.~Wang, R.~A. Popa,
  and I.~Stoica.
\newblock Judgebench: A benchmark for evaluating llm-based judges.
\newblock In \emph{The Thirteenth International Conference on Learning
  Representations (ICLR)}, 2025.
\newblock URL \url{https://openreview.net/forum?id=G0dksFayVq}.

\bibitem[Tang et~al.(2024)Tang, Zhou, Li, Ji, Hou, and Zhang]{tang2024citeeval}
Z.~Tang, K.~Zhou, J.~Li, B.~Ji, J.~Hou, and M.~Zhang.
\newblock L-citeeval: Do long-context models truly leverage context for
  responding?
\newblock \emph{arXiv preprint arXiv:2410.02115}, 2024.

\bibitem[Tian et~al.(2024)Tian, Gao, Zhang, Chen, Fan, Guo, Haas, Ji,
  Krongchon, Li, et~al.]{tian2024scicode}
M.~Tian, L.~Gao, S.~Zhang, X.~Chen, C.~Fan, X.~Guo, R.~Haas, P.~Ji,
  K.~Krongchon, Y.~Li, et~al.
\newblock Scicode: A research coding benchmark curated by scientists.
\newblock \emph{Advances in Neural Information Processing Systems},
  37:\penalty0 30624--30650, 2024.

\bibitem[Treude and Gerosa(2025)]{treude2025developers}
C.~Treude and M.~A. Gerosa.
\newblock How developers interact with ai: A taxonomy of human-ai collaboration
  in software engineering.
\newblock In \emph{2025 IEEE/ACM Second International Conference on AI
  Foundation Models and Software Engineering (Forge)}, pages 236--240. IEEE,
  2025.

\bibitem[Vaghefi et~al.(2023)Vaghefi, Stammbach, Muccione, Bingler, Ni, Kraus,
  Allen, Colesanti-Senni, Wekhof, Schimanski, and
  Leippold]{vaghefi2023chatclimate}
S.~A. Vaghefi, D.~Stammbach, V.~Muccione, J.~Bingler, J.~Ni, M.~Kraus,
  S.~Allen, C.~Colesanti-Senni, T.~Wekhof, T.~Schimanski, and M.~Leippold.
\newblock Chatclimate: Grounding conversational ai in climate science.
\newblock \emph{Communications Earth \& Environment}, 4\penalty0 (1):\penalty0
  480, 2023.

\bibitem[Wang et~al.(2023)Wang, Li, Chen, Zhu, Lin, Zhang, Ren, Sun, and
  Qiu]{wang2023large}
P.~Wang, L.~Li, L.~Chen, D.~Zhu, B.~Lin, Y.~Zhang, Y.~Ren, T.~Sun, and X.~Qiu.
\newblock Large language models are not fair evaluators, 2023.

\bibitem[Wang et~al.(2024)Wang, Li, Chen, Cai, Zhu, Lin, Cao, Kong, Liu, Liu,
  and Sui]{wang-etal-2024-large-language-models-fair}
P.~Wang, L.~Li, L.~Chen, Z.~Cai, D.~Zhu, B.~Lin, Y.~Cao, L.~Kong, Q.~Liu,
  T.~Liu, and Z.~Sui.
\newblock Large language models are not fair evaluators.
\newblock In L.-W. Ku, A.~Martins, and V.~Srikumar, editors, \emph{Proceedings
  of the 62nd Annual Meeting of the Association for Computational Linguistics
  (Volume 1: Long Papers)}, pages 9440--9450, Bangkok, Thailand, Aug. 2024.
  Association for Computational Linguistics.
\newblock \doi{10.18653/v1/2024.acl-long.511}.
\newblock URL \url{https://aclanthology.org/2024.acl-long.511/}.

\bibitem[Wang et~al.(2025)Wang, Tan, Jin, Xiong, Hu, Zhang, Lu, and
  Zhang]{wang-etal-2025-medcite}
X.~Wang, M.~Tan, Q.~Jin, G.~Xiong, Y.~Hu, A.~Zhang, Z.~Lu, and M.~Zhang.
\newblock {M}ed{C}ite: Can language models generate verifiable text for
  medicine?
\newblock In W.~Che, J.~Nabende, E.~Shutova, and M.~T. Pilehvar, editors,
  \emph{Findings of the Association for Computational Linguistics: ACL 2025},
  pages 18891--18913, Vienna, Austria, July 2025. Association for Computational
  Linguistics.
\newblock ISBN 979-8-89176-256-5.
\newblock \doi{10.18653/v1/2025.findings-acl.967}.
\newblock URL \url{https://aclanthology.org/2025.findings-acl.967/}.

\bibitem[Wei et~al.(2025)Wei, Wen, Qiao, Sun, and Ma]{weirocketeval}
T.~Wei, W.~Wen, R.~Qiao, X.~Sun, and J.~Ma.
\newblock Rocketeval: Efficient automated llm evaluation via grading checklist.
\newblock In \emph{The Thirteenth International Conference on Learning
  Representations}, 2025.

\bibitem[Wu et~al.(2025)Wu, Wu, Wei, Zhang, Casasola, Nguyen, Riantawan, Shi,
  Ho, and Zou]{wu2025automated}
K.~Wu, E.~Wu, K.~Wei, A.~Zhang, A.~Casasola, T.~Nguyen, S.~Riantawan, P.~Shi,
  D.~Ho, and J.~Zou.
\newblock An automated framework for assessing how well llms cite relevant
  medical references.
\newblock \emph{Nature Communications}, 16\penalty0 (1):\penalty0 3615, 2025.

\bibitem[Xie et~al.(2025)Xie, Weng, Zhu, Shen, Huang, Lin, Zhou, Mao, Yang,
  Yang, Wu, and Zhang]{Xie2025HowFAC}
Q.~Xie, Y.~Weng, M.~Zhu, F.~Shen, S.~Huang, Z.~Lin, J.~Zhou, Z.~Mao, Z.~Yang,
  L.~Yang, J.~Wu, and Y.~Zhang.
\newblock How far are ai scientists from changing the world?
\newblock \emph{ArXiv}, abs/2507.23276, 2025.
\newblock URL \url{https://api.semanticscholar.org/CorpusId:280401075}.

\bibitem[Yue et~al.(2024)Yue, Ni, Zhang, Zheng, Liu, Zhang, Stevens, Jiang,
  Ren, Sun, Wei, Yu, Yuan, Yang, Ding, Tang, Zadanyi, Liang, Liu, Li, Chen,
  Cheng, Liu, Qiu, Huang, Xie, Ong, Zhang, Wang, and Rui]{mmmu}
X.~Yue, Y.~Ni, K.~Zhang, T.~Zheng, R.~Liu, G.~Zhang, S.~Stevens, D.~Jiang,
  W.~Ren, Y.~Sun, C.~Wei, B.~Yu, R.~Yuan, R.~Yang, M.~Ding, J.~Tang,
  Z.~Zadanyi, X.~Liang, Q.~Liu, W.~Li, E.~Chen, P.~Cheng, Y.~Liu, X.~Qiu,
  X.~Huang, X.~Xie, Y.-S. Ong, Y.~Zhang, J.~Wang, and Y.~Rui.
\newblock Mmmu: A massive multi-discipline multimodal understanding and
  reasoning benchmark for expert agi, 2024.

\bibitem[Zeng et~al.(2023)Zeng, Yu, Gao, Meng, Goyal, and
  Chen]{zeng2023evaluating}
Z.~Zeng, J.~Yu, T.~Gao, Y.~Meng, T.~Goyal, and D.~Chen.
\newblock Evaluating large language models at evaluating instruction following.
\newblock \emph{arXiv preprint arXiv:2310.07641}, 2023.

\bibitem[Zhang et~al.(2025)Zhang, Bai, Lv, Gu, Liu, Zou, Cao, Hou, Dong, Feng,
  and Li]{zhang-etal-2025-longcite}
J.~Zhang, Y.~Bai, X.~Lv, W.~Gu, D.~Liu, M.~Zou, S.~Cao, L.~Hou, Y.~Dong,
  L.~Feng, and J.~Li.
\newblock {L}ong{C}ite: Enabling {LLM}s to generate fine-grained citations in
  long-context {QA}.
\newblock In W.~Che, J.~Nabende, E.~Shutova, and M.~T. Pilehvar, editors,
  \emph{Findings of the Association for Computational Linguistics: ACL 2025},
  pages 5098--5122, Vienna, Austria, July 2025. Association for Computational
  Linguistics.
\newblock ISBN 979-8-89176-256-5.
\newblock \doi{10.18653/v1/2025.findings-acl.264}.
\newblock URL \url{https://aclanthology.org/2025.findings-acl.264/}.

\bibitem[Zhang et~al.(2024)Zhang, Ding, Lv, Wang, Yin, Zhang, Yu, Wang, Li,
  Xiang, Xiang, Wang, Qin, Zhang, Zhang, Cui, Xu, Chen, Fan, Xing, and
  Chen]{Zhang2024ScientificLLA}
Q.~Zhang, K.~Ding, T.~Lv, X.~Wang, Q.~Yin, Y.~Zhang, J.~Yu, Y.~Wang, X.~Li,
  Z.~Xiang, Z.~Xiang, Z.~Wang, M.~Qin, M.~Zhang, J.~Zhang, J.~Cui, R.~Xu,
  H.~Chen, X.~Fan, H.~Xing, and H.~Chen.
\newblock Scientific large language models: A survey on biological \& chemical
  domains.
\newblock \emph{ACM Computing Surveys}, 57:\penalty0 1 -- 38, 2024.
\newblock URL \url{https://api.semanticscholar.org/CorpusId:267301629}.

\bibitem[Zheng et~al.(2023)Zheng, Chiang, Sheng, Zhuang, Wu, Zhuang, Lin, Li,
  Li, Xing, et~al.]{zheng2023judging}
L.~Zheng, W.-L. Chiang, Y.~Sheng, S.~Zhuang, Z.~Wu, Y.~Zhuang, Z.~Lin, Z.~Li,
  D.~Li, E.~Xing, et~al.
\newblock Judging llm-as-a-judge with mt-bench and chatbot arena.
\newblock \emph{Advances in Neural Information Processing Systems}, 36, 2023.

\bibitem[Zhu et~al.(2025)Zhu, Wang, and Wang]{zhujudgelm}
L.~Zhu, X.~Wang, and X.~Wang.
\newblock Judgelm: Fine-tuned large language models are scalable judges.
\newblock In \emph{The Thirteenth International Conference on Learning
  Representations (ICLR)}, 2025.

\end{thebibliography}
